\documentclass[a4paper,10pt]{article}
\usepackage{hyperref}
\pdfoutput=1
\usepackage{geometry}
\usepackage{url}
\usepackage[font=small,labelfont=bf]{caption}
\usepackage{natbib}

\usepackage{placeins}
\usepackage{xcolor}
\hypersetup{
    colorlinks,
    linkcolor={red!50!black},
    citecolor={blue!50!black},
    urlcolor={blue!80!black}
}
\usepackage{tablefootnote}
\usepackage[english]{babel}
\usepackage[utf8]{inputenc}
\usepackage{amsmath}
\usepackage{graphicx}
\usepackage[colorinlistoftodos]{todonotes}
\usepackage{multicol}
\usepackage{caption}
\usepackage{subcaption}
\usepackage{algorithm}
\usepackage[noend]{algpseudocode}

\newenvironment{packed_itemize}
{
	\begin{itemize}
	  \setlength{\itemsep}{1pt}
	  \setlength{\parskip}{0pt}
	  \setlength{\parsep}{0pt}
}
{
	\end{itemize}
}

\usepackage[strict]{changepage}

\usepackage{framed}

\definecolor{formalshade}{rgb}{0.95,0.95,1}

\newenvironment{formal}{%
  \MakeFramed{\advance\hsize-\width\FrameRestore}%
  \noindent\hspace{-4.55pt}% disable indenting first paragraph
  \begin{adjustwidth}{}{7pt}%
  \vspace{2pt}\vspace{2pt}%
}
{%
  \vspace{2pt}\end{adjustwidth}\endMakeFramed%
}
\title{Unsupervised Learning from Continuous Video in a Scalable Predictive Recurrent Network}

\author{
  Filip Piekniewski\thanks{Corresponding author.}\\
  \texttt{filip@piekniewski.info}
  \and
  Patryk Laurent\\
  \texttt{laurent@braincorporation.com}
  \and
  Csaba Petre\\
  \texttt{csaba.petre@gmail.com}
  \and
  Micah Richert\\
  \texttt{richert@braincorporation.com}
  \and
  Dimitry Fisher\\
  \texttt{fisher@braincorporation.com}
  \and
  Todd L. Hylton\\
  \texttt{thylton@eng.ucsd.edu}
}
\begin{document}
\maketitle

\begin{abstract}
\noindent 
Understanding visual reality involves acquiring common-sense knowledge about countless regularities in the visual world, e.g., how illumination alters the appearance of objects in a scene, and how motion changes their apparent spatial relationship. These regularities are hard to label for training supervised machine learning algorithms; consequently, algorithms need to learn these regularities from the real world in an unsupervised way. We present a novel network meta-architecture that can learn world dynamics from raw, continuous video. The components of this network can be implemented using any algorithm that possesses three key capabilities: prediction of a signal over time, reduction of signal dimensionality  (compression), and the ability to use supplementary contextual information to inform the prediction. The presented architecture is highly-parallelized and scalable, and is implemented using localized connectivity, processing, and learning. We demonstrate an implementation of this architecture where the components are built from multi-layer perceptrons. We apply the implementation to create a system capable of stable and robust visual tracking of objects as seen by a moving camera. Results show performance on par with or exceeding state-of-the-art tracking algorithms. The tracker can be trained in either fully supervised or unsupervised-then-briefly-supervised regimes. Success of the briefly-supervised regime suggests that the unsupervised portion of the model extracts useful information about visual reality. The results suggest a new class of AI algorithms that uniquely combine prediction and scalability in a way that makes them suitable for learning from and --- and eventually acting within --- the real world.
\end{abstract}

%\begin{multicols}{2}

\section{Introduction}
We are interested in creating practical vision systems that can power front-ends for applications like autonomous robots, self driving cars, or intelligent security systems.  By ``practical,'' we mean vision systems that can function in real time, in the real world, be scalable, and perform well when faced with challenging visual conditions. 
Objects and scenes are subject to numerous temporal and contextual effects, including changes in lighting, shadows, reflections, motion blur, and partial occlusions.  
Although conceptually the physical dynamics that, e.g., underlie an object rolling into and out of a shadow are relatively simple, inferring these effects from the pixel values projected onto a camera is a non-trivial task, as illustrated in Figure~\ref{fig:greenball}.  Building systems that can learn the \emph{common-sense} knowledge to reliably process such scenes has long remained an unsolved problem.  Nearly 30 years ago, Hans Moravec pointed out the problem of common-sense perception for artificial intelligence (AI) systems \citep{moravec1988mind} in the following paradox:

\begin{formal}
\emph{It is comparatively easy to make computers exhibit adult level performance on intelligence tests or playing checkers, and difficult or impossible to give them the skills of a one-year-old when it comes to perception and mobility.} 
\end{formal}

\begin{figure}[th!]
\centering
\includegraphics[width=0.75\textwidth,trim={0cm 8.5cm 8cm 3cm},clip]{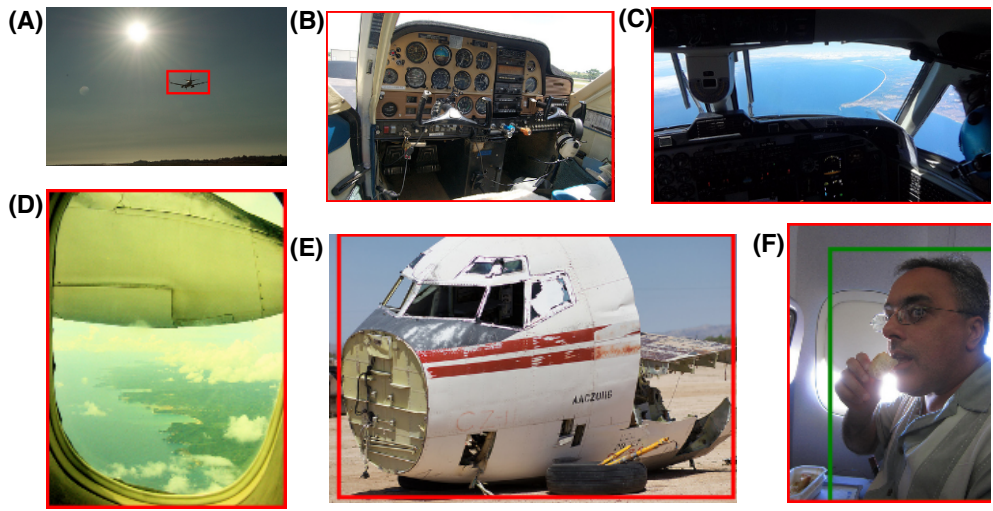}
\caption{\textbf{Direct Mapping from Images to the Human Category Label ``Airplane''.}   In the ImageNet competition, deep networks are tasked with mapping images like these to human category labels. The images in this figure are labeled with the category ``airplane''. In our opinion, ImageNet is not a good benchmark task for general AI learning systems, because correct generalized classification of these images requires much more than just visual inspection. Humans access additional contextual, dynamical and cultural knowledge when identifying each image in a general way: e.g., (A) airplanes fly, rather than just remain on the ground, (B) airplanes have complex controls in their cockpits, (C, D) views from an airplane's windows include the ground, (E) what airplanes look like when they are in pieces, (F) the fact that humans can order and enjoy snacks on an airplane.  Contemporary, feedforward deep convolutional nets trained offline on ImageNet are unable to abstract out and learn these facts. It is, however, possible for a system with many parameters to memorize these images and textures in a non-generalizable way. (Images from ILSVRC 2014).}
\label{fig:bad_task}
\end{figure}

\noindent To-date the issues raised by Moravec have not been directly tackled or successfully addressed by either machine learning or robotics researchers, despite the massive progress in general computing.  In the case of machine learning research, the focus on games has continued -- from checkers, to chess \citep{campbell2002deep}, Jeopardy \citep{lewis2012game}, and most recently, Go \citep{silver2016mastering} and Atari games \citep{mnih2013playing}.  Benchmarks such as ImageNet \citep{imagenet_cvpr09} or CIFAR-10 \citep{krizhevsky2009learning} have moved research in the direction of perception, but as we point out in later sections, high-level adult-human categorization of objects does not facilitate the kind of \emph{common sense} knowledge learning implied by Moravec's paradox.  ImageNet, despite its use of real-world images, is heavily skewed by human cognitive and perceptual category labels, which are based not just on visual information, but on myriad contextual, functional and cultural factors, as shown in Figure~\ref{fig:bad_task}.  In light of Moravec's paradox, ImageNet and similar benchmarks are not the kind of tasks we expect from a one year-old. In robotics, at the same time, efforts have been focused on building algorithms that work very well in restricted environments  \citep[c.f., Universe of Discourse][]{lighthill1973artificial}. The results of the recent DARPA robotics challenge clearly show the extent to which problems of perception, manipulation, and mobility remain very challenging in unrestricted environments \citep{atkesonhappened}. See also \cite{sigaud2016towards} for thorough review of of challenges for modern machine learning architectures in robotics.

Our approach, described in this paper, involves rethinking both the nature of the benchmark task and the construction of machine learning components used for perception.  In the spirit of Moravec's paradox, we shied away from adult-level tasks like image categorization and opted instead for visual object tracking as a task that is challenging, practical, yet one that might be expected of a one-year-old. We call the architecture that we are developing based on these ideas the Predictive Vision Model (PVM). As we will claim below, the architecture has  the properties necessary to discover \emph{common-sense} dynamical information from the world that is currently missing from machine learning approaches.

This paper is organized as follows: first we provide background on the machine learning problem in computer vision and identify the shortcomings in current deep learning techniques (Section~\ref{sec:background}). Next we present our novel  architecture and a neural network implementation of it that attempts to overcome these shortcomings (Section~\ref{sec:general}). We show results from three experiments related to visual object tracking (Section~\ref{sec:experiments}). We discuss the results in Section~\ref{sec:discussion}. Next we provide concluding remarks and future research directions in Section~\ref{sec:conclusions}.

\section{Background}
\label{sec:background}
\subsection{History: Neuroscientific origins of machine learning in computer vision}

Modern machine learning approaches to computer vision originate primarily from the idea of the Neocognitron \citep{fukushima1980neocognitron}.  The Neocognitron in turn was largely inspired by a series of seminal studies in the biology of the mammalian visual system. Studies by David Hubel and Torsten Wiesel on cat primary visual cortex asserted that vision arises as a primarily feedforward hierarchy of alternating functions, namely, feature construction and invariance-generating pooling \citep{hubel1959receptive,hubel1962receptive}\footnote{Fundamental to the design of the Neocognitron was the idea of ``simple'' and ``complex'' cells.  Simple cells were orientation-selective cells with potent responses to a fixed stimulus bar at particular orientations and eccentricities -- called ``simple cell'' receptive fields -- in layer 4 of primary visual cortex, and orientation selective cells with broader invariant response to spatial shifts -- called "complex cell" receptive fields -- in layers 2 and 3.}. The Neocognitron inspired many subsequent models including the H-max model \citep{riesenhuber1999hierarchical,serre2005object}, and LeNet \citep{lecun1995comparison}. 

The recent renaissance of connectionist approaches to vision under the moniker \emph{deep learning} \citep{lecun2015deep} is a direct continuation of the above mentioned work, but in the context of significantly increased computing power and certain optimized initialization and regularization techniques that allow the models to contain many more layers than in previous work \citep{krizhevsky2012imagenet,srivastava2014dropout}. In many cases, including modern deep learning, modelers were inspired by the findings of ``complex'' cells and built in layers composed of max operators that work on a weight shared feature map, generating translational invariance. 

Attempts thus far to incorporate other biologically-inspired concepts like invariances based on slowly-changing features \citep{foldiak1991learning,wiskott2002slow} have met with limited success.  The potential benefits of other concepts from biology and cognitive science, like prediction, recurrent connectivity, top-down feedback, and highly local learning, have yet to be realized.

\begin{figure}[ht]
\centering
\includegraphics[width=0.9\textwidth]{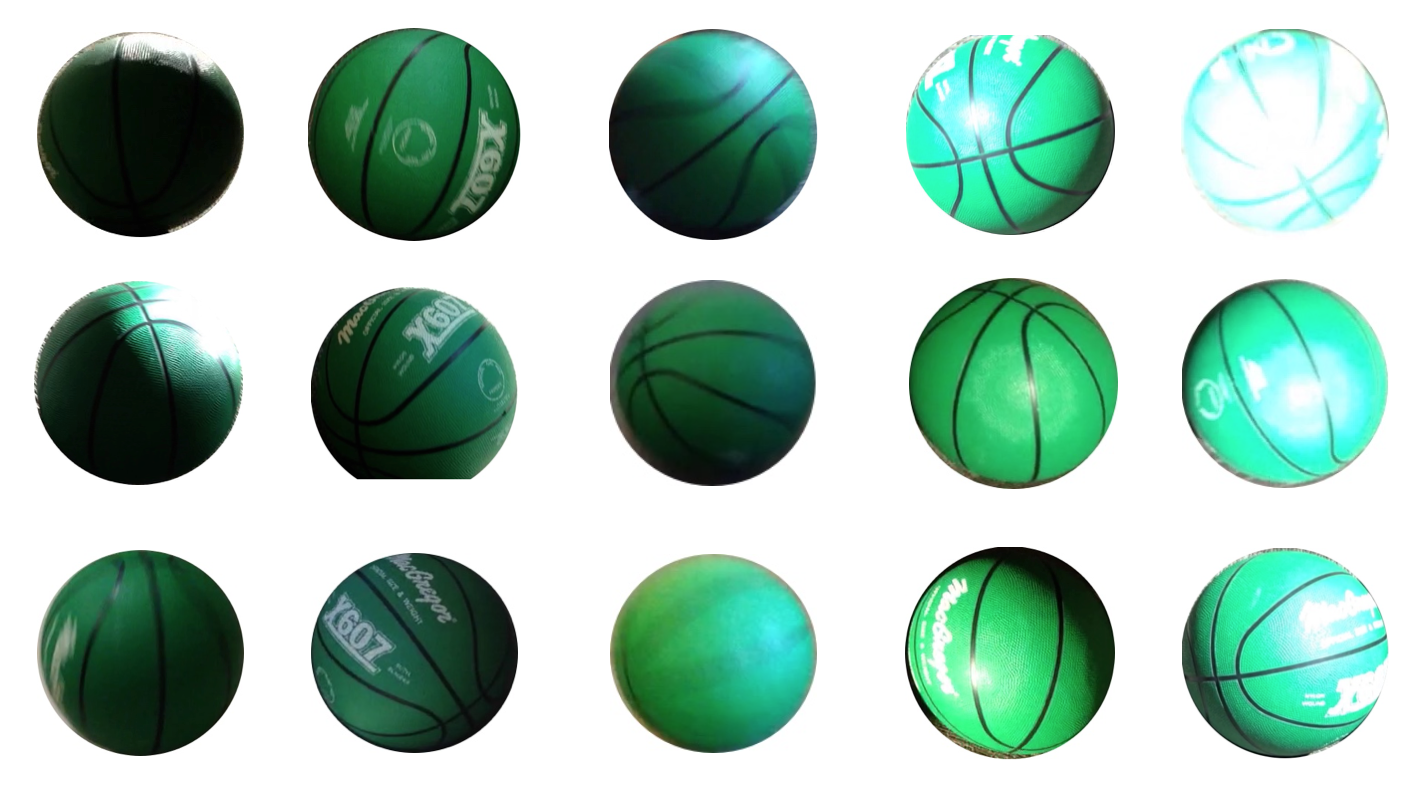}
\caption{\label{fig:greenball} \textbf{The Same Object at Different Times and in Different Contexts}. An artificial vision system is challenged when tasked with simply classifying whether images contain a green basketball. This figure shows crops of a green basketball from video as it rolled around on the ground in a naturally lit room with an open window.  Notice that this single object looks markedly different in different frames. The way an objects appears depends strongly on ambient contextual effects (e.g., lighting, background, shadows), temporal context (e.g., motion, rotation), and even the dynamics of the camera. The colors in these images vary in hue from blue-green to yellow-green. Shadow and specular reflections alter the apparent shape of the ball.  Motion blur momentarily obscures features like texture. Purely feedforward algorithms cannot use either context or time to classify these static images; rather, they are ``forced'' to find ways to group together clusters of visual space that are meaningfully linked through physical transformations.  Because of the problems of high-dimensionality described in the text, feedforward algorithms are highly unlikely to discover the true general decomposition of these spatial and temporal effects and apply them across objects.}
\end{figure}

\subsection{Image classification and the problem of generalization}
\label{sec:classification-generalization}
To motivate our work, this section reviews the typical machine vision problem that deep learning techniques are primarily used to solve today --- image classification.

In image classification, the stated problem is to associate some vector of high dimensional data points with some lower dimensional representation --- usually an image to a class label --- in a way that generalizes to new data. For example, an image could be labeled (classified) as whether or not it depicts a catfish. In the case of binary classification, the high dimensional image vectors have to be projected onto a set composed of 0 and 1 (indicating whether the sample belongs to a category or not) or to a [0, 1] segment if fractional category membership is allowed. 

The set of data samples used to train the system is called a ``training set'' and is assumed to be a fair (i.e., unbiased) representation of the true, unknown distribution of the data.  Under this assumption, it is often concluded that given enough sample data, the correct values of metaparameters, and enough training time, the solution found by the training procedure of the machine learning algorithm will generalize to correctly classify new samples. In other words, the algorithm will be able to output the correct associations for input vectors it has not seen before. This statement should be taken with caution, as even though there are important theoretical results such as the universal approximation theorem \citep{gybenko1989approximation,hornik1989multilayer}, whether a good solution can be achieved in practice is by no means guaranteed and subject to many conditions \citep{domingos2012few}.

In cases where the training set is not sufficiently representative of the actual data distribution, the machine learning algorithm may form a non-general solution, which gives good performance on the training set, but very poor performance on actual new data. In fact, with enough parameters and after sufficiently long training time, even relatively simple algorithms (e.g., a multilayer perceptron trained using backpropagation of error) are likely to return a solution that will always perform very well on the training set but poorly on new data.  Seemingly paradoxically, performance on the new data may even decrease with more training on the training set. This phenomenon --- in which the algorithm performs worse on new data with more training on the training set --- is called overfitting (see Chapter 1 in \cite{Bishop:2006:PRM:1162264} as well as \cite{hawkins2004problem,domingos2012few}. 

Overfitting does not just depend on the specifics of the algorithm, but also very much depends on the properties of the data: Consider for comparison purposes the easily-generated cases of artificial datasets that exist on low dimensional manifolds but are submerged into very high-dimensional spaces --- for example, a Gaussian cloud spanning only a few dimensions, or a set of image classes that reliably differ only by a simple pixel tag (e.g., a single pixel in the corner of the image that contains all the information needed to classify the image). These two cases can be handled very well by most machine learning systems as the boundary separating the classes is rather regular and can be inferred from a relatively low number of training samples.  However, even artificially generated problems such as the two-spiral set \citep{lang1988learning} can be very hard to separate. Given its origin in complex physical reality (e.g., see Figure~\ref{fig:greenball}), visual data may span hundreds of dimensions and may have fractal-like regions of much greater complexity than spirals. The number of samples of data needed for supervised machine learning algorithms to correctly separate such regions (i.e., in a way that generalizes) can easily exceed any number that would be practical to collect and label. 
 
Deep convolutional nets attempt to address these problems by exploiting assumptions of translational invariance of textures and features, which help to reduce the dimensionality of the visual signal being processed as well as the number of parameters needed in the model.  These assumptions make the learning problem more tractable, and make it possible to train networks with more layers. However, even deep convolutional networks suffer from numerous problems leading to their inability to generalize as humans would \citep{nguyen2015deep, papernot2016limitations, goodfellow2014explaining}. Importantly, as observed in \cite{goodfellow2014explaining} and \cite{szegedy2013intriguing}, these mis-classifications (``adversarial examples'') are not just isolated cases, but entire subspaces of inputs that persist across multiple deep learning models trained on a variety of data. Furthermore in \citep{kurakin2016adversarial} researchers are even able to pass adversarial examples through printout and re-acquisition of image via a camera. These deficiencies likely stem at least in part from the fact that translational symmetry is merely an approximation to a group of real symmetries of the data manifolds.  In the case of these high dimensional problems, reducing the number of parameters to mitigate over-fitting is ultimately futile: in actuality, far more parameters are needed to adequately capture the complexity of real visual scenes.  However, with more parameters comes the need for strong constraints. Translational invariance is just one way of constraining parameters in the end-to-end training paradigm but, as noted above, is not sufficient. There are many other transformations and changes in visual appearance that preserve the identity of objects which could serve to constrain parameters in a predictive paradigm.  These transformations are difficult for people to recognize and to code as priors.

In addition, direct end-to-end mappings using deep networks are still very technically challenging to configure for successful learning. The large number of layers makes credit assignment difficult, a problem known as the vanishing gradient \citep{glorot2010understanding, hochreiter1998vanishing}. Recent techniques such as orthogonal weight initialization or pre-training with restricted Boltzmann machines \citep{hinton2006fast} partially help to deal with this issue, but have not solved it entirely. Although some very deep networks have been reported \citep{1512.03385}, these are often built using advanced ``residual nets'' techniques where the true minimal depth of the network --- expressed as the shortest path from input to output layer --- is not actually larger than more conventional deep learning networks. 

Despite these challenges, deep networks can learn representations that are useful for a number applications.  These representations are semantically relevant and have been used for image search, content-aware advertising, content filtering and so on. However, these transformations are insufficient for general scene understanding and vision for autonomous action. Below we summarize the problems we believe need to be addressed to achieve robust visual perception:
\begin{packed_itemize}
\item Visual data consists of very high dimensional input, and exists on complex manifolds embedded in high dimensional spaces. Convolutional approaches cannot replicate the complexity of the general manifold.
\item The vanishing gradient in end-to-end training paradigms is only partially addressed by convolutional feature maps, residual networks, orthogonal initialization, or other such methods.
\item Good generalization is exceedingly unlikely with the amount of labeled data available; there is a much higher probability of ``memorizing'' textures which leads to peculiar failure modes \citep{nguyen2015deep, papernot2016limitations,goodfellow2014explaining}
\end{packed_itemize}

\subsection{Beyond the Neocognitron}
\label{sec:missing-ingredients}
As we've argued in previous sections, we believe that current algorithms based largely on Neocognitron-type architectures \citep{fukushima1980neocognitron} only partially address the problems of dimensionality reduction, learning, and robust visual perception.  Close examinations of conflicting bodies of research within neuroscience, along with our own requirements for robust robotics vision algorithms, compelled us to take a different approach that abandoned Neocognitron-based designs entirely. Evidence in neuroscience has accrued that the predominantly feedforward, hierarchical picture presented by Hubel and Wiesel is greatly simplified. First of all, anatomical studies of the neocortex point to substantial feedback connectivity \citep{douglas2007recurrent}, which significantly affects even the responses in peripheral parts of the primary visual cortex \citep{girard2001feedforward,hupe1998cortical}.  The functional taxonomy into simple and complex cells is not nearly as clear cut as was originally claimed \citep{fournier2011adaptation}, as there are numerous modulation effects of spatial and temporal context on any receptive field \citep{angelucci2006contribution}. Aside from this, the majority of visual learning in early childhood is clearly unsupervised, which is not addressed by current models very well.  Finally, there is no evidence in support of weight sharing in the cortex, which is a common optimization in deep learning networks.

Given the above, we opted to investigate building a model that could encompass ubiquitous feedback, and learn in a primarily unsupervised way. Inspired by ideas of brains as predictive systems \citep{clark2013whatever, rao1999predictive, palm2012prediction, vondrickanticipating,ranzato2014video} and the requirement to develop computer vision algorithms that could scale to real world robotics applications, we designed an architecture that:

\begin{packed_itemize}
\item Uses unsupervised learning (mediated by prediction) of regularities in the visual world
\item Incorporates recurrence/feedback to facilitate discovery of regularities across multiple temporal scales
\item Features a multiscale hierarchical organization, to capture regularities spanning multiple spatial scales
\item Uses local learning to overcome the vanishing gradient problem that makes credit assignment difficult in deep hierarchies
\item Builds using composable components, in a scalable manner
\item Builds out of simple components (associative memory units), without being wedded to any particular implementation of those components
\end{packed_itemize}

\subsection{Selecting the benchmark task}

Selecting appropriate benchmark tasks is a critical and often overlooked step in building a new machine learning architecture \citep{torralba2011unbiased}. On one hand, selecting a benchmark that incorporates aspects of practical tasks helps to ensure that the new algorithm may find applications. On the other hand, focusing on a single task may hamper generality and robustness of the new algorithm, as inevitably, priors about the data will leak into its design.  

Benchmarks in machine learning are useful to the extent to which they allow for the creation of new algorithms, but their value diminishes once the performance of any algorithm approaches the theoretical maximum \citep{wagstaff2012machine} (typically indicative of ``meta-overfitting''\footnote{Another way of saying that multiple tests on the entire dataset, optimization of meta-parameters by hundreds of researchers over many years deprives further results obtained on such sets from any statistical significance, see page 1108 in \cite{gabbay2011philosophy}}). Benchmarks that have been available for a long time (e.g., MNIST \cite{lecun1998mnist}) are increasingly subject to the multiple comparisons problem\footnote{Also known as the ``look-elsewhere effect'' or ``knowledge leakage'' in Machine Learning \citep{elkan2012evaluating}, this stems effectively from the fact that once an algorithm was trained and evaluated on a test set, then that same test set can no longer be used to evaluate subsequent modifications to the algorithm without losing statistical significance.}. Results obtained from such benchmarks lack statistical significance and in the worst case lead to data ``dredging'' \citep{gabbay2011philosophy, giles1997presenting, smith2002data, elkan2012evaluating}. For a publicized incident, see \cite{Markoff:2015aa}. Finally, another problem is that various pressures for novelty and publication impact among scientists and researchers can create the harmful illusion that if a new algorithm does not outperform state of the art in one of the well established benchmarks then it is not worthy of further study. 

With these constraints and cautionary notes in mind we sought a task/benchmark for PVM that would:
\begin{packed_itemize}
\item Be aligned with a problem that is currently unsolved and challenging for current approaches
\item Be natural to formulate as an online, real-time task
\item Have several existing state-of-the-art, general algorithms to compare against
\item Be a first step in a progression towards more complex benchmarks 
\item Allow for easy gathering of data
\item Allow experimenters to test hypotheses about what the system ``knows'' given its training
\item Be relatively low-level on the scale of perceptual and cognitive abilities (i.e., not overly influenced by high-level human cognition, categories, or ambiguities.)
\end{packed_itemize}
A task that adequately meets the above requirements is visual object tracking. 

\subsection{Visual object tracking}
\label{sec:visual-object-tracking}
Visual object tracking is a well known, challenging, yet easily definable task in machine vision --- making it suitable for a benchmark. Even though substantial research has gone into making robust tracking systems for particular object types, general object tracking continues to be a very difficult, unsolved problem (see Section~\ref{sec:existing-trackers}).  A number of academic and commercial algorithms have been designed for tracking particular objects characterized by distinct features, such as histograms of oriented gradients \citep{dalal2005histograms} or the use of keypoints as in the SIFT algorithm \citep{lowe2004distinctive}. General tracking of arbitrary objects is, however, much more difficult.  In this section we describe tracking and the benefit of using tracking as a benchmark task in more detail.

Briefly, the goal of visual object tracking is to provide a bounding box around the object of interest in a continuous sequence of images. The object may be occluded or out of view in some number of the frames (the algorithm should then report the absence of the target). As the sequence of images continues, the appearance of the object may change substantially as it undergoes transformations and illumination changes. The online aspect of this task means that the sequence of images is presented to the algorithm and results should be returned after each new frame (prior to the next one). Consequently the algorithm may have access to the history of the sequence but does not have access to the entire sequence upfront, much like an agent acting in a physical environment. Thus, tracking enforces online processing, offers data with rich temporal correlations, and can be unambiguously benchmarked. 

As a benchmark task, visual object tracking naturally fulfills our requirement that ML systems should use temporal structure: typically datasets are video clips showing moving objects, often with hand-labeled bounding boxes outlining some target object. Visual object tracking is also of practical importance and can be used in multiple applications including: approaching objects, avoiding obstacles, reaching for objects, sorting objects (e.g., on conveyors), and navigating towards or around objects.   Unlike image classification, visual object tracking does not rely on human-created category labels\footnote{A tracker does not need to know the category membership or identity of the object it is tracking, neither its cultural context , but it should know how it transforms in physical environment.}. Tracking is not subject to artificially inflated scores, as has been the case when algorithms have expert knowledge that most humans do not have (e.g., super-human recognition of a sub-species, see breeds of dogs in ImageNet \cite{imagenet_cvpr09}). To avoid fixation on one task, tracking can be expanded into a series of increasingly complex tasks, like target pursuit and object manipulation tasks. When benchmark scores become high, the next more complex task should be used\footnote{An older benchmark can still be used for development if it is of practical importance, but it should not be used for research on general AI anymore.}.

\subsubsection{Existing state-of-the-art tracking algorithms}

\label{sec:existing-trackers}

There are various existing solutions to visual object tracking \citep[for a comprehensive list, see references in][]{WuLimYang13}.  Some tracker algorithms are highly engineered for a particular class of objects (e.g., human faces), whereas others attempt to be more general.  General trackers typically have a brief \emph{priming phase} in which the tracker is ``instructed'' to track the object of interest. The priming phase typically employs some sort of machine learning technique that operates on one or more static images and searches for a particular distribution of image features. Details of the trackers included for comparison in our tracking benchmark are provided in Section~\ref{sec:trackers-used}.

Some trackers, like TLD \citep{kalal2012tracking}, incorporate motion information during learning or priming and make use of the resulting independent, but unreliable, estimate of where an object is.  In these usages, motion is computed explicitly by comparing the difference between many successive frames.  The resulting information is used, e.g., to improve the training of a feature-based classifier. However, much of the information present in the motion itself --- i.e., the dynamics of motion for the object of interest, and the resulting image transformations --- is ignored. Rather, motion is simply used as a means to identify the set of static image features belonging to the object, visible from different vantage points across time. Thus, in state-of-the-art trackers, although motion helps extract information about objects, motion itself is not directly incorporated into the learning part of the algorithm.

Besides motion information, there is a lot of other valuable information in visual signals that is also not used but could be enormously informative as to the position of the object and its expected appearance.
This includes the global scene context (which includes shadow distribution, source of illumination, positions of occluders, etc.) as well as temporal context (e.g., the fact that the object was recently occluded or is motion blurred because of a sudden shake of the camera). These valuable pieces of information are not used by state-of-the-art trackers since they are very hard to realize in a rigidly coded algorithm.

There are several recent papers employing deep learning for online object tracking \citep{hong2015online, wang2015visual}. In summary these approaches use pre-trained layers of convolutional features as a front-end to a classifier that then generates a heatmap by sliding the classifier over the entire image. These features were trained on ImageNet and for reasons we elaborate on in section \ref{sec:classification-generalization} cannot represent any temporal context of the scene. Such an approach can provide good results in certain limited cases such as relatively small benchmarks \citep{WuLimYang13}, but are effectively equivalent to existing trackers where hand-engineered features are replaced with a convolutional net front-end.  

In summary, existing trackers track appearances of objects but know little about the physical reality in which those objects exist. We believe that this is a fundamental problem that can only be addressed by learning the basic physical properties of reality as presented to a visual sensor.

\subsubsection{Existing visual object tracking benchmarks} 

There are multiple existing visual object tracking benchmarks, e.g.:
\begin{packed_itemize}
\item Geiger, Lenz \& Urtasun (KITTI; 2012) \citep{Geiger2013IJRR}
\item Wu, Lim \& Yang (2013) \citep{WuLimYang13}
\item Oron, Bar-Hillel, Avidan (2015) \citep{oron2015real}
\item VOT challenge \url{http://www.votchallenge.net}
\end{packed_itemize}

\noindent The above-listed benchmarks are all suitable for classical tracking algorithms, but are not adequate for developing the kinds of algorithm we propose. Both KITTI \citep{Geiger2013IJRR} and the benchmark developed by \cite{oron2015real} are focused on automotive applications, which contain primarily one type of complex, human-defined object --- cars. \cite{WuLimYang13} is a general benchmark composed out of smaller datasets released along with tracking algorithms and contains many short clips with a large variety of different objects, consequently each object is represented by a very small amount of data.  Similarly, the VOT challenge aims at (quoted from their website): ``single-object short-term trackers that do not apply pre-learned models of object appearance (model-free)''. Instead, PVM focuses on building models of not just the target object but of broader visual reality. In addition, the algorithms we are developing require more data in the form of labeled video than just a single ``priming'' frame, which is the methodology used for most other trackers. We also require the captured video to be unadulterated -- that is, it should include challenging conditions including lighting changes, shadows, reflections, backlighting, and lens flares.  Finally, it is important that the continuous, unadulterated video be shot from many different and changing viewing angles \citep[addressed to some extent in the benchmark by][]{oron2015real}). Having investigated the benchmarks above we decided to create a moderately sized dataset that would fit our needs better (see Section~\ref{sec:pvm_dataset}).

\begin{figure}[th!]
\centering
\includegraphics[width=0.4\textwidth,trim={0cm 1cm 0cm 0.5cm},clip]{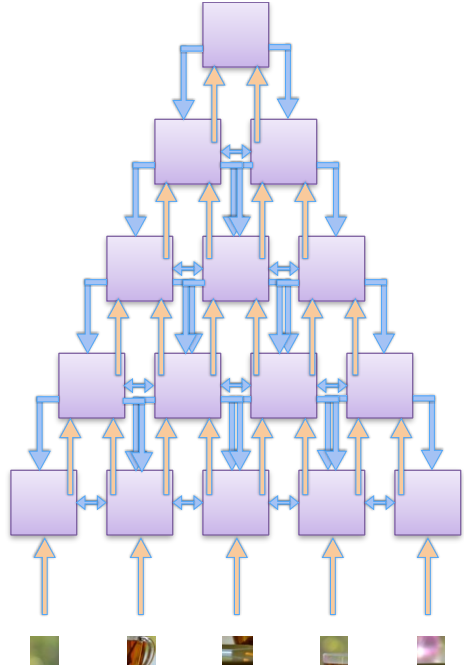}
\caption{\label{fig:pvm_hierarchy} 
\textbf{Meta-Architectural Scale View of the PVM Circuit.}  This figure shows an example of PVM processing units and their inter-connectivity. PVM is a recurrent, predictive model meta-architecture of units (purple boxes) with near-neighbor (lateral) and top-down (feedback) connectivity. Sensory information, here an image from a video stream, is divided into tiles and presented at the input of the network (bottom).  At each level, units process their input signal and send dimensionality-reduced signal to the next level of the hierarchy, creating the inputs at that level (orange arrows). To aid in local prediction, each unit also receives contextual information from both lateral and superior units (blue arrows). Feedback connectivity can also be configured to skip across levels. E.g., in the simulations presented here, context from the top-most is also fed back to all units in the network (connections not shown for clarity).
}
\end{figure}

\section{General Methods}
\label{sec:general}

In this section we will describe the architecture we have developed, and the methodology we used for training and testing it. The architecture is abstract in that it is not strongly dependent on the specifics of the algorithm within each unit -- hence we use the term \emph{meta-architecture}. The architecture design is driven by two basic requirements: (1) it must learn and operate online with real world data (e.g. video), and (2) it must scale simply and efficiently (in hardware and software).  The general notion of prediction of the future local signal is a powerful training paradigm that satisfies the first constraint. The architecture is composed of interconnected processing units,  each of which learns to locally predict the signal it receives, using prior values of the signal and contextual information from neighboring units. Scalability is met by several properties of the meta-architecture: it is composed of a large number of identical processing units, shows good convergence time in the number of training parameters, has highly local and stable learning, comes with high availability of training data thanks to its unsupervised training paradigm, and has the ability to function without a global clock. We elaborate on these aspects of scalability in Section~\ref{sec:scalability}.

Although PVM does allow for deep hierarchies, it is substantially different from existing deep convolutional networks. PVM is recurrent, with ubiquitous feedback connectivity. Indeed, in our implementations, the majority of the input to each unit is recurrent input from lateral and top-down feedback connections\footnote{Which is what appears to be observed in biological cortex as well \citep{douglas2007recurrent}.}. Recurrent feedback allows PVM to discover regularities in time as well as effects of spatial and temporal context. Each PVM unit broadcasts a compressed representation of its prediction as signal input or context to other, connected units (see Figures~\ref{fig:pvm_hierarchy} and ~\ref{fig:pvm_unit}).  In the implementation described below, units are mostly connected to their neighbors. This connectivity reflects the fact that in vision applications the nearby spatial information is the most relevant.  Nothing in the architecture, however, prevents non-neighboring, non-hierarchical connectivity which might be advantageous in other applications.

PVM is predisposed to extract common-sense regularities from the data, unlike models that have to build direct mappings between 2D visual images and high-level concepts.  As we discussed previously in Section~\ref{sec:classification-generalization}, such direct mappings are complex and fraught with numerous discontinuities.   

\subsection{PVM architectural details}

The Predictive Vision Model (PVM) is a collection of associative memory units connected into a pyramid-like hierarchy\footnote{The pyramid-like structure or hierarchy are not a hard requirements of the PVM design. Other structures are possible.} as shown in Figure~\ref{fig:pvm_hierarchy}.  Each unit in this architecture can be thought of as an abstract processing unit serving as an associative memory. Each PVM unit:
\begin{packed_itemize}
\item receives a ``primary'' signal of moderate dimensionality (on the order of 100-d)
\item builds an association from context and an input signal to future values of that input signal
\item predicts the next value of the signal based on the learned association
\item creates an intermediate, compressed (reduced dimensionality) representation of the prediction suitable for transmitting to connected units\footnote{All implementations of PVM units must output a reduced dimensionality signal for the architecture to scale, see Section~\ref{sec:scalability}.}
\item has an optional ``readout'' layer that can be trained, via supervision, to serve as a task-related output (see Figure~\ref{fig:pvm_unit}). These readouts will be later used in constructing a heatmap that will serve as output of the tracker.
\end{packed_itemize}

Refer to Table \ref{table:comp} for a brief comparison highlighting the differences between PVM and existing deep learning approaches.

\begin{table}[!ht]
\resizebox{\textwidth}{!}{%
\begin{tabular}{ |l|p{5cm}|p{5cm}| }
\hline
 & \textbf{Deep Learning} & \textbf{Predictive Vision Model} \\
\hline
\textbf{Information Flow} & Feedforward\tablefootnote{The use of recurrence in deep learning is typically limited to LSTM based layers. In PVM feedback is global and naturally wires between any part of the model. Refer to section \ref{sec:comparison} for more detailed comparison to similar approached in deep learning} & Extensive Feedback  \\
\hline
\textbf{Computation} & End-to-end error propagation & Local prediction  \\
\hline
\textbf{Spatio-Temporal Context} & Rarely used 
\tablefootnote{There are recent attempts \citep{misra2016cross} to share parts of two models for two different tasks, an approach that might be viewed as use of lateral context or co-prediction. In PVM we built this feature deep into the fabric of the architecture, such that it becomes essential part of it. We discuss in section \ref{sec:comparison} several attempts, mainly from the deep learning community to incorporate such features} & Top-Down/Integrated  \\
\hline
\textbf{Data} &  Shuffled frames/vectors & Continuous video \\
\hline
\textbf{Advantages of Neuromorphic Impl.} &  Small & Large\tablefootnote{We elaborate in section \ref{sec:path_neuromorphic} on why we believe this is the case. Notably there are hardware implementations of deep networks, but they more resemble a vectorized CPU's than a uniform neuromorphic fabric.} \\
\hline

\textbf{Learning} &  Offline, primarily Supervised & Online, primarily Unsupervised  \\
\hline
\textbf{Example Tasks} & Human-centric Classification &  Visually guided behavior such as object tracking  \\
\hline
\end{tabular}}
\caption{High level comparison of existing approaches in deep learning and the proposed PVM architecture.}\label{table:comp}
\end{table}

\subsubsection{PVM unit implementation details}

\label{sec:unit_details}

\begin{figure}[ht]
\centering
\includegraphics[width=0.7\textwidth,trim={0cm 6cm 0cm 6cm},clip]{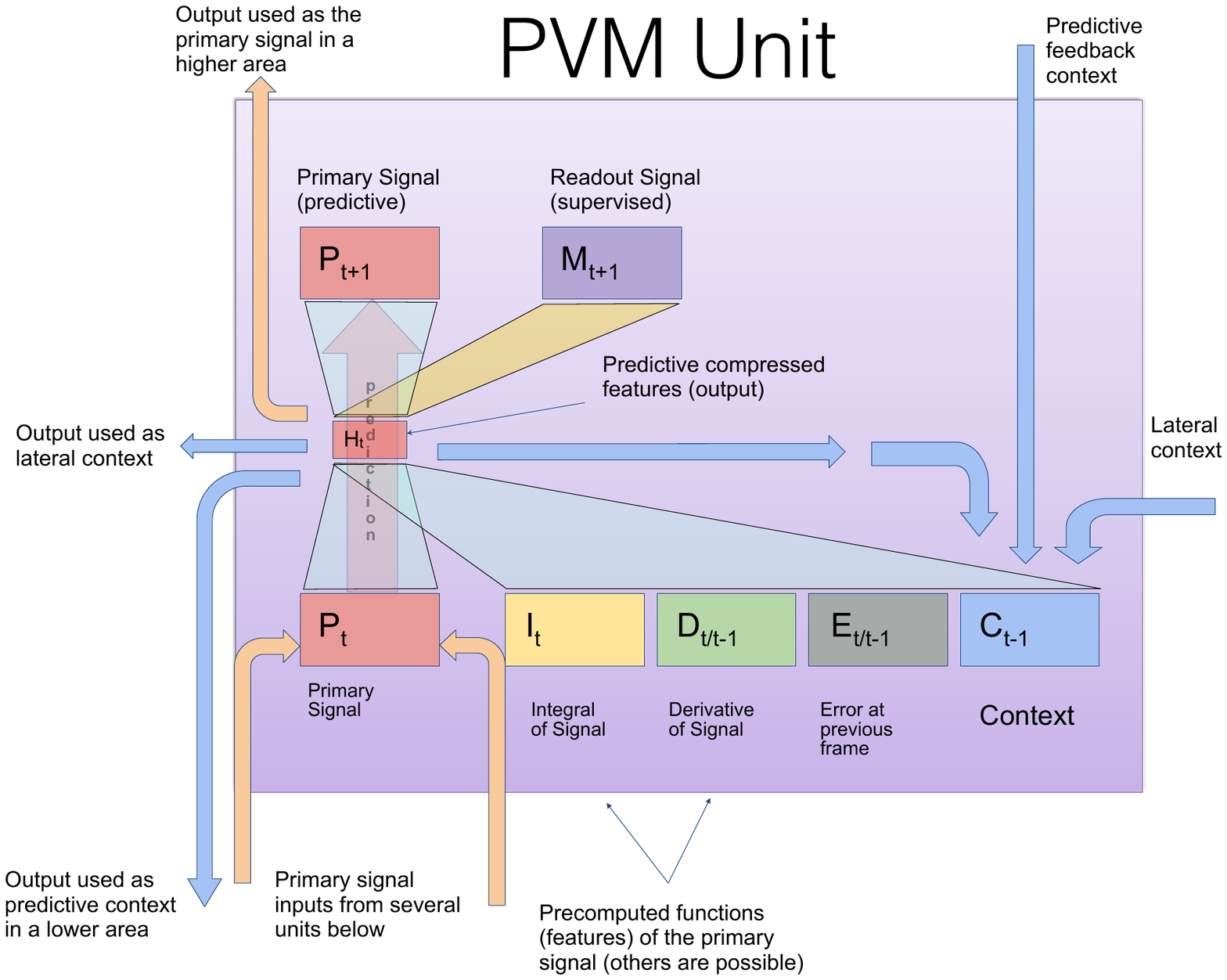}
\caption{\label{fig:pvm_unit} \textbf{Detail View of a Single PVM Unit.} In the present implementation, a PVM unit is built around a multilayer perceptron (MLP). The two inputs to the PVM unit are the primary signal (ascending), and the context inputs (lateral and descending). The two inputs are concatenated with locally computed features and fed into the MLP (see Table~\ref{table:summary}). The output of a PVM unit is the compressed representation from the hidden layer of the MLP, which according to PVM design assumptions is of strictly lower dimension than the primary signal. In the first layer of PVM units, the primary input signal (to use and predict) consists of a subset of tiles from the image.  At higher stages of the PVM hierarchy, the primary input signal (to use and predict) consists of concatenated compressed representations from a lower level.  $P_t$: Primary Input Signal; $P_{t+1}$: Primary Predicted Signal, self-supervised; $M_{t+1}$: Readout Signal, supervised (i.e., used to construct the tracker heatmap). }
\end{figure}

Refer to Figure \ref{fig:pvm_unit} for a diagram of a PVM unit and Algorithm \ref{algorith:pvm_unit} for pseudo-code. In the current implementation, each unit is built around a multilayer perceptron (MLP) with sigmoid activation neurons \citep{rumelhart1985learning} trained with stochastic gradient descent online (without shuffling).  The task of each unit is to predict the future of this primary signal, i.e., to generate a prediction $P^*_{t+1}$ that approximates the true $P_{t+1}$. Dimensionality reduction is achieved through the ``bottleneck'' provided by an MLP with a smaller hidden layer than the input and output layers --- similarly to a denoising auto-encoder \citep{vincent2008extracting}. 

The input signal for each MLP consists of three parts.  The first part of the input is the primary signal $P_t$, which comes from the video input or from the output of lower level units.   The second part of the unit's input consists of context inputs from lateral and superior units (as well as the top unit, in this particular tracking application).  In addition, pre-computed features of the primary signal are added as a third part of the input.  See Table \ref{table:summary} for details.

\makeatletter
\def\BState{\State\hskip-\ALG@thistlm}
\makeatother

\begin{algorithm}
\caption{PVM unit (runs in parallel for all units, synchronised by the barrier)}\label{algorith:pvm_unit}
\begin{algorithmic}[1]
\Procedure{PVM\_unit\_run}{barrier}
\State memory1 \Comment{Represents associative memory for the predictive part}
\State memory2 \Comment{Represents associative memory for the task readout part}
\State \Comment{memory1 and memory2 can have shared parts}
\While {True}
\State $\textbf{Synchronize}(\textit{barrier})$
\State $\textit{input} \gets \text{Concatenate}(\textit{signal, precomputed\_features, context})$
\State $\textit{signal\_prediction} \gets \text{Make\_Prediction}(\textit{memory1, input})$ \Comment{memory activations are a function of the input}
\State $\textit{readout\_prediction} \gets \text{Make\_Prediction}(\textit{memory2, input})$ 
\State $\textit{output\_signal} \gets \text{Get\_Compressed\_Representation}(\textit{memory1})$ \Comment{These are functions of the input}
\State $\text{Publish}(\textit{output\_signal})$ \Comment{Make output available to everyone else}
\State $\text{Publish}(\textit{readout\_prediction})$ \Comment{Used to create the tracker readout}

\State $\textbf{Synchronize}(\textit{barrier})$
\State $\textit{signal} \gets \text{Collect\_Current\_Signal}()$ \Comment{Actual next value of the primary signal} 
\State $\textit{precomputed\_features} \gets \text{Precompute\_Features}(\textit{signal})$ \Comment{Integral, derivative etc.}
\State $\textit{readout} \gets \text{Collect\_Current\_Readout}()$ \Comment{Supervising  signal for the task}
\State $\textit{p\_error} \gets \text{Calculate\_Prediction\_Error}(\textit{signal\_prediction, signal})$
\State $\textit{r\_error} \gets \text{Calculate\_Readout\_Error}(\textit{readout\_prediction, readout})$
\State $\text{Train}(\textit{memory1, p\_error})$
\State $\text{Train}(\textit{memory2, r\_error})$
\State $\textbf{Synchronize}(\textit{barrier})$
\State $\textit{context} \gets \text{collect\_current\_context}()$
\EndWhile
\EndProcedure
\end{algorithmic}
\end{algorithm}
As noted above, PVM utilizes local learning and is currently implemented with ``shallow'' three-layer perceptrons. Even though the entire model can be considered deep --- the model reported here has six layers of hierarchy, see below --- we did not need to use of any of the recently developed deep learning techniques like convolution \citep{lecun1995comparison}, half-rectified linear units \citep{nair2010rectified}, dropout \citep{srivastava2014dropout} and so forth. It is worth emphasizing that there is nothing specific about this particular unit-level implementation\footnote{We have preliminarily explored  sending gradient information among units via their feedback connections as a part of their context (data not reported).  This would make the meta-architecture dependent on gradient-based learning methods. However, our preliminary findings suggest that sending gradient information as part of context is not crucial for the network to function.} and other algorithms can used as the associative memory as long as they produce a compressed representation of the prediction, e.g., Boltzmann machines \citep{ackley1985learning} or spiking networks capable of compression \citep[e.g.,][]{August:1999hc, levy2005interpreting}.

\begin{figure}[ht]
\centering
\includegraphics[angle=90, width=0.7 \textwidth,trim={0.5cm 0cm 0.5cm 0cm},clip]{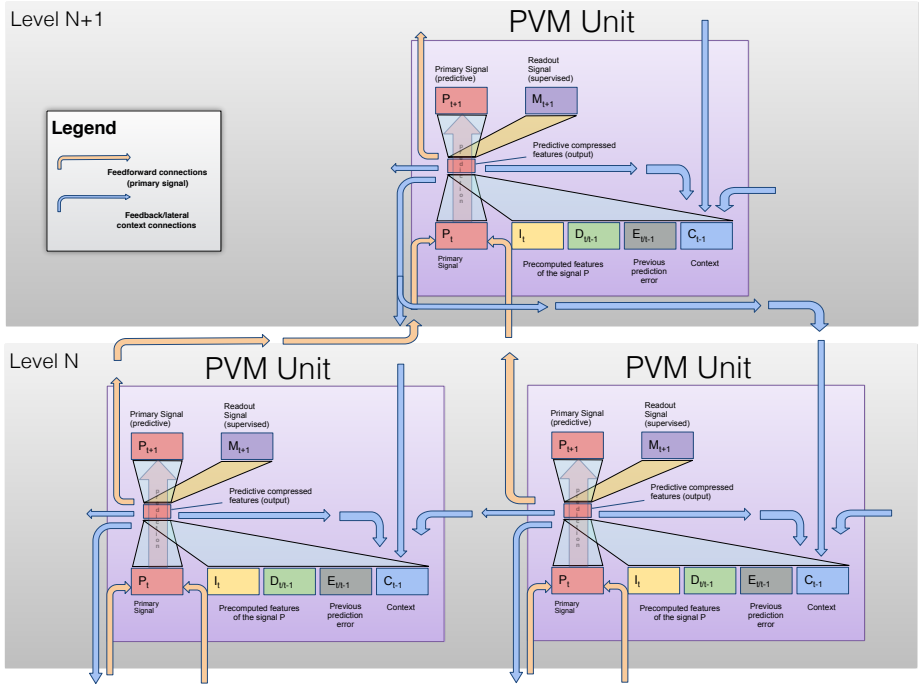}
\caption{\label{fig:pvm_circuit} \textbf{PVM Local Circuit}. Several units in the lower level of the hierarchy provide their compressed predictions to a unit in the upper level. Each unit sends its own predictions up (as primary signal) as well as down and lateral (as context). }
\end{figure}

The compressed representations from the units are concatenated and serve as the primary input signal for units in the next level of the hierarchy (see Figure~\ref{fig:pvm_circuit}). In addition those same compressed representations serve as \emph{context} for their lateral neighbors as well as units lower in the hierarchy (via recurrent feedback connections). The context serves as an additional input alongside the primary signal that allows each unit to make better predictions and as a memory akin to that found in the Simple Recurrent Network \citep{elman1990finding}.  We further elaborate on a surprising relationship between PVM and Simple Recurrent Networks in Section~\ref{sec:alternative-context-interpretation}.

Feedback and lateral connections are a critical part of this architecture.  They can be liberally wired everywhere in the hierarchy. For example, units at the top-most level of the hierarchy can send their feedback projections all the way down to the input layer. Although the locality of physical events and economy of connectivity are good reasons for predominantly local connections, long range projections may be helpful for efficiently capturing global illumination effects. Importantly, because each PVM unit is locally self-supervised by the primary signal, amplifying loops in the feedback connectivity are inherently unstable as they dissipate as soon as they start corrupting the predictions. Thus the feedback is self-stabilizing.

\begin{table}[!ht]
\resizebox{\textwidth}{!}{%
\begin{tabular}{ |l|c|c|c| }
\hline
Layer or Sub-layer & Symbol & Number of neurons in PVM unit & Definition \\
\hline
\multicolumn{4}{ |c| }{\textbf{Inputs}} \\
\hline
Signal & $P_t$ & variable $\sim 100$  &  Fan-in from inferior layer or raw video tile \\
\hline
Integral & $I_t$ & same as $P_t$  & $\tau * I_{t-1} + (1 - \tau) P_t$ \\
\hline
Derivative & $D_{t/t-1}$ & same as $P_t$ & $1/2 + (P_t - P_{t-1}) / 2$ \\
\hline
Previous Prediction Error & $E_{t/t-1}$ & same as $P_t$ & $1/2 + (P^*_{t} - P_{t}) / 2$  \\
\hline
Context  & $C_{t-1}$ & variable $\sim 200$ & $  \textrm{concat[} H_t ... \textrm{]}  $ from Hidden of self/lateral/superior/topmost units \\
\hline
\multicolumn{4}{ |c| }{\textbf{Output}} \\
\hline
Hidden  & $H$ & 49 & $\sigma(W_h \cdot [P_t ;  D_{t/t-1} ; I_t ; E_{t/t-1} ; C_t] )$  \\
\hline
\multicolumn{4}{ |c| }{\textbf{Predictions}} \\
\hline
Predicted Signal & $P^*_{t+1}$ & same as $P_t$  & $ \sigma(W_p \cdot H_t) $ (not sent out) \\
\hline
Tracking Task  & $M*_{t+1}$ & variable $1\sim 256$ & $  \sigma(W_m \cdot H_t) $ To heat map  \\
\hline
\end{tabular}}
\caption{\textbf{Summary Table of a PVM Unit.} Each PVM unit consists of a three layer MLP with sigmoid activation neurons. Here $P^*_t$ and $P^*_{t+1}$ are the predictions of the primary signal at times t and t+1, whereas $P_t$ is the actual primary signal at time t. Values provided are for the models presented in this paper; other values are possible. Refer to the published source code (\url{http://www.github.com/braincorp/PVM}) for the exact way these values are obtained. }\label{table:summary}
\end{table} 

\subsubsection{Specific architectural parameter values}

The particular models described in this paper had the following sizes and values:

\begin{packed_itemize}
\item Six layers of hierarchy, first $16\times 16$, second $8\times 8$, third $4\times 4$, fourth $3\times 3$, fifth $2\times 2$, sixth $1$ PVM units, respectively.
\item The input image was $96\times 96$ RGB pixels, chopped into $6\times 6$ RGB tiles (i.e., $6\times 6 \times 3=108$ inputs), and each tile was fed into a corresponding PVM unit in the first layer.
\item The compressed representation formed by each PVM unit (the MLP's hidden layer) was $7\times 7=49$ sigmoid neurons.
\item Each PVM unit in the higher levels of the hierarchy received a concatenation of 4 ($2\times 2$) compressed representations from underlying PVM units.
\item Each PVM unit received lateral context from its four nearest neighbors (except for those at the edge, which received context from fewer).
\item If a PVM unit sends signal to a PVM unit in a higher level of the hierarchy, then the upper one sends feedback context back down to the lower one.
\item The top-most PVM unit sends context back to all units in the hierarchy.
\item The readout heatmap resolution was $16\times 16$ pixels for all layers (entire layer) except for fifth where it was $18\times 18$ pixels. Consequently, the readout layer $M_{t+1}$ had the following sizes for each PVM unit: $1\times 1$ pixel for PVM units in first layer, $2\times 2$ pixels in the second, $4\times 4$ pixels in the third, $8\times 8$ pixels in the fourth, $6\times 6$ pixels in the fifth and $16\times 16$ pixels in the sixth.
\item All the parameters used in the models presented here can be found in the file \url{https://github.com/braincorp/PVM/blob/master/PVM_models/model_zoo/experiment1_green_ball.json}.
\end{packed_itemize}

The parameters presented here generated models that we could train in reasonable time (several weeks) given our multicore implementation and available compute hardware (Amazon Web Services \emph{c4.8xlarge} instance or equivalent). Although these values are important for reproducing the current results, they are not the only possible values. Some experimentation has shown that the model is not very sensitive to most of these values (unpublished results). The parameter to which the PVM is the most sensitive is the size of the hidden layer $H$ (i.e., the PVM unit's output), because varying it also changes the sizes of the input to afferent units as well as the size of the context (with fan-in kept constant). Increasing the hidden layer size would result in a supralinear (quadratic) increase in the overall size of the model and training time, both in terms of number of required training trials and total execution time.

\subsubsection{Tracker readout}

\begin{figure}[ht]
\centering
\includegraphics[ width=\textwidth]{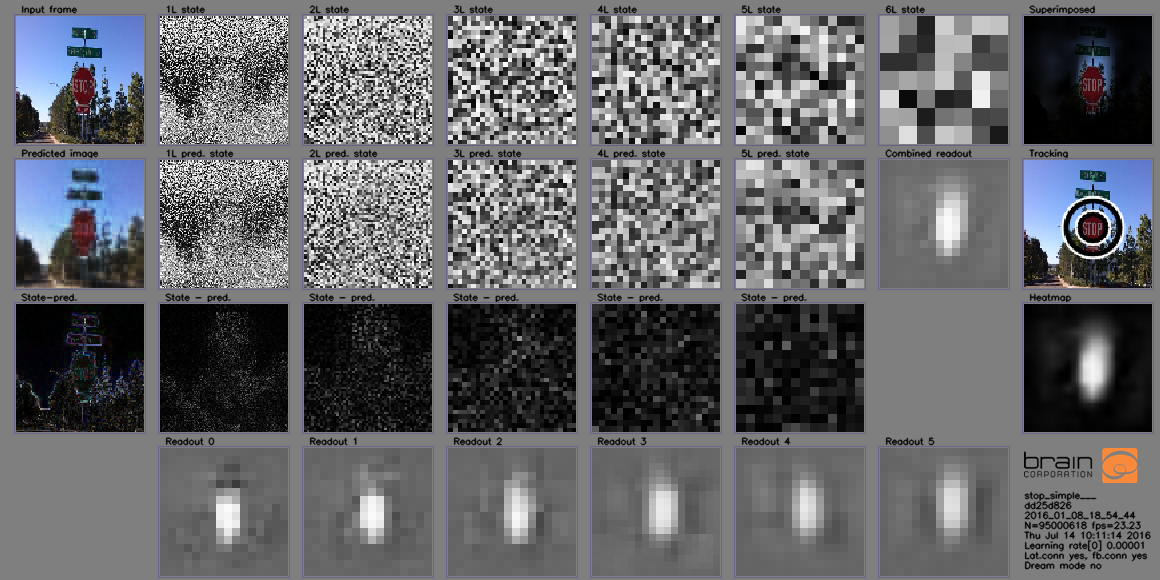}
\caption{\textbf{PVM Running in Our Software Framework}. Shown is one of the models discussed in this paper with six layers of hierarchy. Top row from left to right: input frame, internal compressed representations of subsequent layers (columns 2-7), visualization of tracking heatmap superimposed on the input image (last column). Second row from the top: predictions generated by each layer (columns 1-6), averaged tracking heatmap, tracking bulls-eye. Third row: differences between signal and predictions for each layer (columns 1-6), averaged tracking heatmap scaled to [0-255] (last column), last row: individual per-layer tracking heatmaps. Note the target (stop sign) position is recovered in all the layers including the last one which is 7 perceptron layers away from the input. }\label{fig:pvm_screen}
\end{figure}

\begin{figure}[ht]
\centering
\includegraphics[angle=0, width=0.9\textwidth,trim={0cm 0cm 0 0cm},clip]{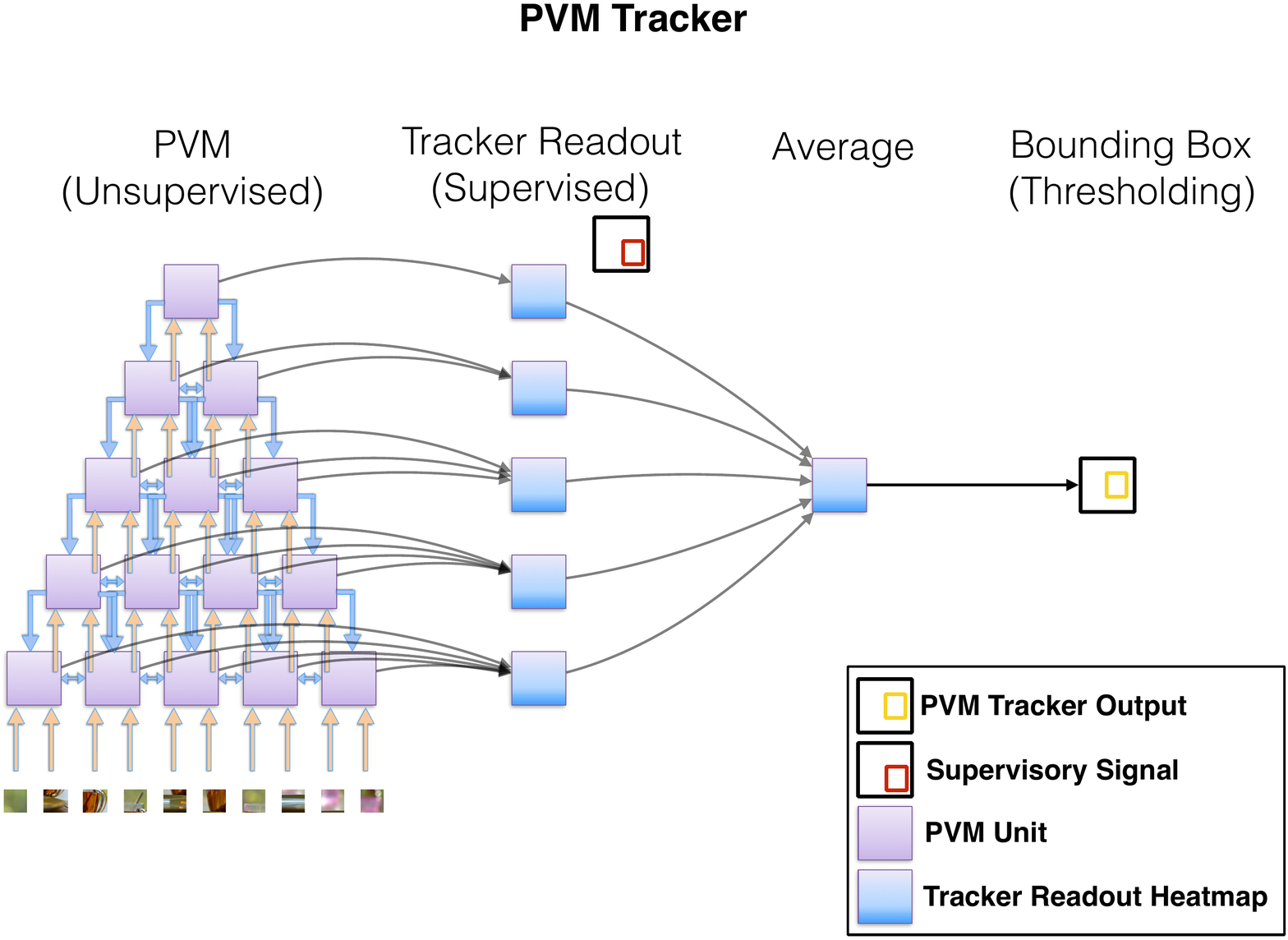}
\caption{\label{fig:pvm_tracker} \textbf{PVM Tracker Design}. The unsupervised PVM network was used as a tracker by training its units' readout layers, $M_{t+1}$, to generate a heatmap at each level of the hierarchy.  These readouts were averaged and thresholded to generate a bounding box suitable for comparing to other trackers (or to human ground-truth labeling).}
\end{figure}

The tracker readout refers to the estimate of the location of the tracked object by the PVM tracker. The tracker readout is computed by averaging the output $M_{t+1}$ of multiple PVM units (Figures~\ref{fig:pvm_screen} and \ref{fig:pvm_tracker}).  This readout is trained via supervised learning on the training dataset, see Section~\ref{sec:training-readout} for more information.

Because the model has many layers and substantial recurrence, we experimented with evaluating the model by allowing it to settle for several steps on each frame of video before estimating the target location and processing the next frame. We found that allowing the model to settle for a few steps on each frame enhanced the tracking performance slightly\footnote{By at most 3 percent.}. The results presented below were obtained with the model allowed to settle for 4 frames. The algorithm to convert the readout into a bounding box is minimal (see Algorithm \ref{algorith:bbox}): it involves identifying the peak of the heatmap, the contour around it, and includes logic to handle the odd case e.g., when the detected peak is outside the traced heatmap contour. The algorithm uses only the heatmaps generated by the PVM --- i.e., it does not include any particular logic/priors about the tracking dataset. The bounding box extraction algorithm was kept simple and fixed over our research period to avoid injecting additional priors into it and artificially inflating our tracking results\footnote{There are likely numerous ways to improve the results by tweaking the final classifier, or even just adjusting the threshold value, but our primary focus was on building the machine learning architecture and demonstrating its usefulness on an important task, rather than on building the best visual tracker.}. The heatmaps from each layer of the hierarchy are averaged and an algorithm to detect deviations from the mean value was employed to make a decision as to whether there is a peak and to estimate the counter around it. Algorithm~\ref{algorith:bbox} has a threshold parameter for detection that was chosen without systematic tuning and is constant for all the results shown in this document. Any reference to PVM tracking or detection threshold refers to this parameter in the bounding box calculation algorithm.

\makeatletter
\def\BState{\State\hskip-\ALG@thistlm}
\makeatother

\begin{algorithm}
\caption{Bounding box creation}\label{algorith:bbox}
\begin{algorithmic}[1]
\Procedure{get\_bounding\_box}{}
\State $\textit{threshold} \gets 32$ \Comment{Fixed in all our simulations}
\State $\textit{combined\_heatmap} \gets \text{mean}(\textit{per layer heatmaps})$
\State $\textit{peak} \gets \text{argmax}(\textit{combined\_heatmap})$
\State $\textit{max\_val} \gets \text{max}(\textit{combined\_heatmap})$
\State $\textit{med\_val} \gets \text{median}(\textit{combined\_heatmap})$
\If {$\textit{max\_val} > \textit{med\_val} + \textit{threshold }$} 
\State $\textit{cutoff} \gets (\textit{max\_val } - \textit{med\_val})*0.5 + \textit{med\_val}$
\State $\textit{binary} \gets \text{cv2.threshold}(\textit{combined\_heatmap}), \textit{cutoff}) $
\State $\textit{contours} \gets \text{cv2.contours}(\textit{binary})$
\State $\textit{contours\_with\_peak} \gets \text{find }(\textit{contours}, \textit{ peak } \in \textit{contour })$
\State $\Return \textit{ box surrounding largest contour with } \textit{peak}$
\Else
\State $\Return \textit{ empty box }$
\EndIf
\EndProcedure
\end{algorithmic}
\end{algorithm}

\subsubsection{Execution time considerations}
\label{sec:execution-time-scalability}
With its current multicore framework implementation, the entire model (as specified in this work) runs at approximately 20 frames per second on an Amazon Web Services \emph{c4.8xlarge} cloud instance (36 cores at 2.9GHz) or \emph{m4.10xlarge} instance (40 cores at 2.5 GHz). This enables an execution rate of approximately 1M steps per 15 hours of computing, while learning is enabled. The model runs 3-4 times faster with learning disabled, as when its performance on the tracking task is being evaluated.

\subsection{Benchmarking and testing the model} \label{sec:benchmarking}

\subsubsection{Tracker benchmark datasets} \label{sec:pvm_dataset}
\begin{figure}[ht]
\centering
\includegraphics[ width=\textwidth]{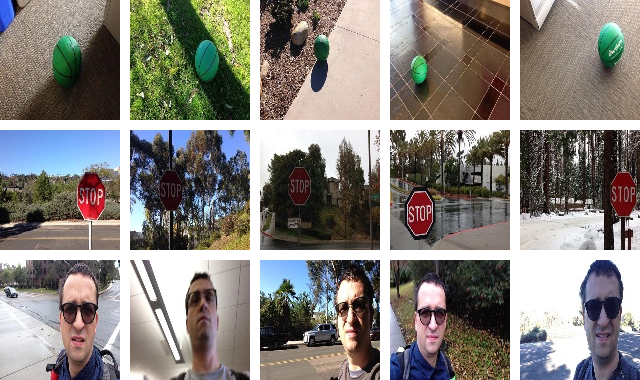}
\caption{\textbf{PVM Dataset Examples}. The dataset contains clips of green basketball rolling on floor/pavement, stop signs seen from 5-15 feet and face (one of the authors) seen from a hand-held camera. The movies are taken in everyday conditions, including varying weather, lighting, etc., using an off-the-shelf camera. The intent was to have general coverage of the kinds of conditions a robot might see in a typical human environment. The entire dataset is available for download at \url{http://pvm.braincorporation.net}.}\label{fig:pvm_dataset}
\end{figure}

Three labeled datasets were created containing movies of: (1) a green basket ball rolling on the floor/pavement, (2) a stop sign viewed from 5-15 feet, and (3) a human face as seen from a handheld camera (see Figure~\ref{fig:pvm_dataset} for a few examples). The datasets contained long portions of continuous video, with the objects entering and exiting the frame of view periodically.  The objects were also subject to numerous lighting condition changes, shadows and specular reflection.

Each data set was divided into a training set and test set. None of the test sets were ever used for training, and none of the reported evaluations were done on the training set. Detailed information on the datasets used for the results in this paper is presented in Table \ref{table_datasets}.  Note that these sets are not very big by ``big data'' standards, nevertheless, the combined test set is larger than the Tracker Benchmark 1.0 dataset \citep{WuLimYang13} used in the literature, which contains 51 movie sequences but only 23,690 frames (the complete PVM dataset is more than $6\times$ larger). Also in terms of raw data, depending on resolution used, the dataset consists of several gigabytes of compressed images, which decompresses into roughly 100GB. 

Finally, we would like to note that it is relatively easy for anyone to create more data. In fact, parts of the test set were recorded after we had already trained some of our models, and we continued to expand the test set as the models were being trained on the training set. The models  generalized very well to the newly acquired test data, see Section~\ref{sec:experiment-1-results}. 

\begin{table}[ht]
\centering
\begin{tabular}{|l|l|l|l|}
\hline
Name & \# sequences& Length (frames) & Approx. duration at 25fps \\ \hline
Green ball training & 5 & 12,764 & 8.5 min \\ \hline
Green ball testing & 29 & 34,791 &  23 min \\ \hline
Stop sign training & 20 & 15,342 & 10 min \\ \hline
Stop sign testing & 30 & 22,240 & 15 min \\ \hline
Face training & 10 & 29,444 & 19.6 min \\ \hline
Face testing & 25 & 34,199 & 23 min \\ \hline
Total training & 35 & 57,550 &  38 min \\ \hline
Total testing & 84 & 91,230 & 60 min\\ \hline
\end{tabular}
\caption{ \textbf{Durations of the PVM Training and Testing Sets.} These values apply to results shown in this paper. The complete data set is published to allow reproduction of the results presented here at \url{http://pvm.braincorporation.net}. Note that the test portions of the set are substantially larger than the training portions. The separation into training and testing is arbitrary (we collected training set first and then kept adding to the testing set).}\label{table_datasets}
\end{table}

\subsubsection{Trackers used for comparison}
\label{sec:trackers-used}
PVM has been tested and compared with the following state of the art trackers that we incorporated into our testing framework:

\begin{packed_itemize}
\item \emph{Struck} - Structured Output Tracking with Kernels \citep{hare2011struck}.
\item \emph{TLD} - Tracking Learning Detection \citep{kalal2012tracking} 
\item \emph{CMT} - Clustering of Static-Adaptive Correspondences for Deformable Object Tracking \citep{Nebehay2015CVPR}
\end{packed_itemize}

Many other trackers exist, but the above three are typically very high performers in several published benchmarks \citep[e.g.,][]{WuLimYang13} and are representative of the current state of the art. We also verified that the performance of STRUCK and TLD agrees with that reported in \citep{WuLimYang13}, confirming that both our tracker implementations and our testing framework are sound.

In addition, we implemented two simple color trackers based on color histograms \citep{swain1992indexing} and Camshift bounding box estimation \citep{bradski1998real}:

\begin{packed_itemize}
\item \emph{UV Tracker} - calculates the histogram of U and V channels in the YUV color space
\item \emph{HS Tracker} - calculates the histogram in Hue and Saturation channels of the HSV color space
\end{packed_itemize}

The performance of color-based trackers is typically indicative of how informative color features are for a given dataset.  Since our dataset contains two subsets of entities for which we expect color to be a very discriminative feature, namely the green basketball set and the stop sign set, we wanted to know if such a simple heuristic would be sufficient to solve the task at a satisfactory level. As the results show in Figure~\ref{fig:Performance_sup}, the color histogram-based trackers listed above performed somewhat better than the state-of-the-art trackers on the green basketball set. However, contrary to our expectations their performance was far from good, and far below the PVM tracker's performance.

We also implemented two control trackers to provide a baseline for performance and to uncover positional biases in the data:

\begin{packed_itemize}
\item \emph{null tracker} - uses the bounding box from the last frame of priming for all tracked frames (assumes the target remains where it was last seen)
\item \emph{center tracker} - uses a bounding box centered on the frame, with 10\% width and height of the frame (assumes the target is in the center).
\end{packed_itemize}

\subsubsection{Tracker performance metrics}
\label{sec:tracker_metrics}
\begin{figure}[ht]
\centering
\includegraphics[width=0.9\textwidth,trim={0cm 6cm 0 4cm},clip]{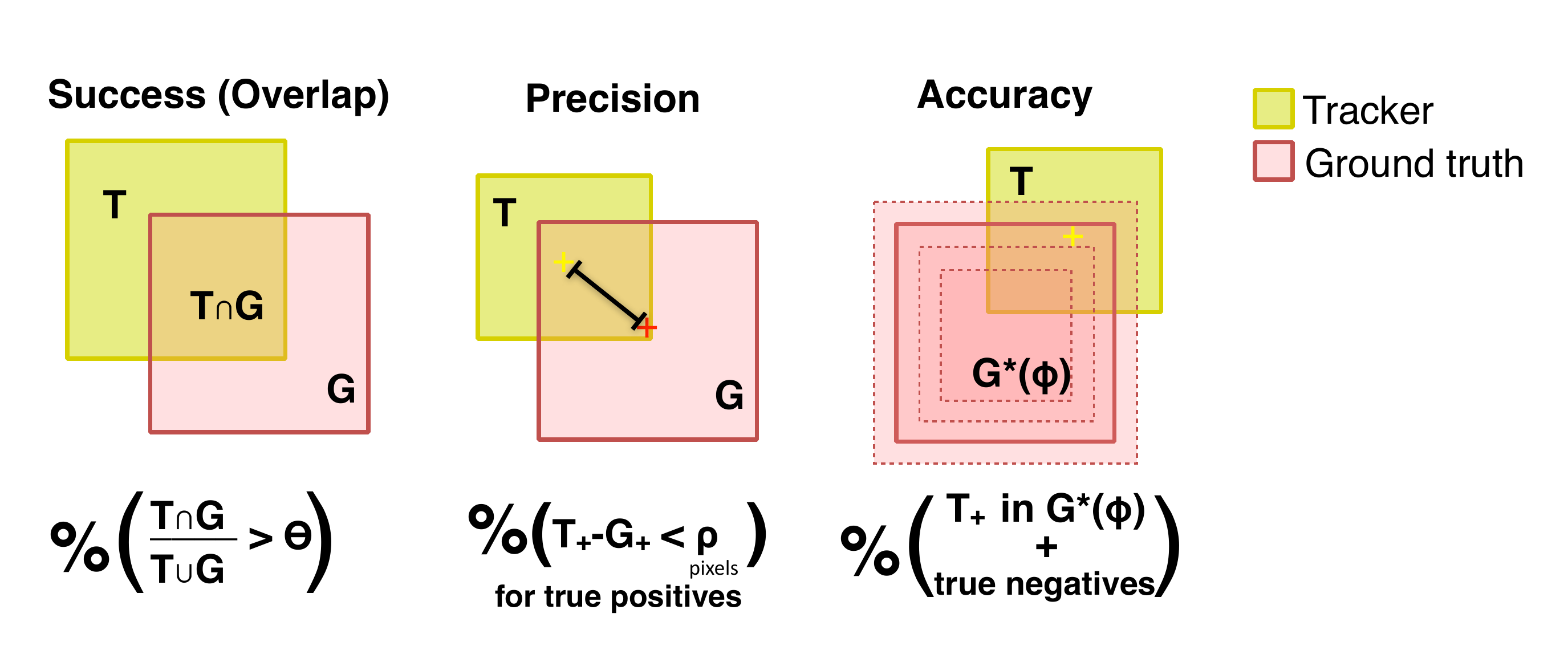}
\caption{\label{fig:trackermetrics}\textbf{Metrics for Evaluating PVM and State-of-the-Art Trackers}.  Three metrics were used to assess the PVM tracker and state-of-the-art trackers.  Success (Overlap) and Precision were previously defined in \cite{WuLimYang13}. We added the measurement of Accuracy which scores a tracker favorably for \emph{true negatives}.  A high score for true negatives is important for many tracking applications: e.g., for a robot pursuing or searching for a target, it is important for the robot to know when the target object really isn't there.}
\end{figure}

Three measures were used to evaluate tracking performance (see Figure \ref{fig:trackermetrics}):
\begin{packed_itemize}
\item \emph{Success (Overlap):} Quantifies the quality of the bounding box.  A Success plot displays the fraction of frames in which the overlap (ratio of area of the intersection to the area of the sum) of the ground truth bounding box and the tracker bounding box is greater than a given argument $\theta$ (from 0 to 1).  The area under the curve quantifies the skewness of this plot, which is greater for better trackers. Success is used in state of the art tracker benchmarking, see \citep{WuLimYang13}.
\item \emph{Precision:} Quantifies the quality of the center of the bounding box \citep[see][]{WuLimYang13}.  Precision measures the fraction of frames in which the distance between the center of the ground truth bounding box and tracked box falls below a given number of pixels, $\rho$.  Precision at $\rho$ = 20 pixels compares performance among trackers.  The threshold of 20 pixels is used in the literature, but the meaning of the result varies as a function of the resolution of the movie frames. Since PVM operates at the relatively low resolution of $96\times 96$ pixels, sampling Precision at  20 pixels is not a particularly informative measure; it merely quantifies the fraction of frames where the tracked bounding box is in the correct general area of the frame. Another disadvantage of Precision is that it is only concerned with true positives and ignores true negatives.  If a tracker always returns a bounding box even when the target is missing, Precision ignores these errors.  These issues led us to develop a measure which is similar to Precision in that it measures the quality of the bounding box center, but also takes into account true negatives; we call it Accuracy (see below).
\item \emph{Accuracy:}  A new metric introduced by us that quantifies tracking with regards to both true positives and true negatives in a resolution-independent manner. Accuracy measures the fraction of frames in which the center of tracked box lies inside the ground truth bounding box plus the fraction of frames in which the target is absent and the tracker also reports that is is absent. To reflect how close the centers are to the ground truth, the ground truth bounding box is scaled by a parameter $\phi$ to generate a curve ($\phi$ = 1.0 means original ground truth bounding box).  Thus, Accuracy is qualitatively similar to Precision but, because it operates in the units of the ground truth bounding box size, has the benefit of being resolution independent. Further, it properly evaluates trackers for times that the target is absent by incorporating the true negative rate into the score.
\end{packed_itemize}

\subsubsection{A note on resolution}

All of the results presented here were accomplished at the native resolution of the PVMs we have developed ($96\times 96$).  All videos were scaled to this resolution for all trackers.

\section{Experiments}
\label{sec:experiments}

Experiments 1 and 2 tested hypotheses about the kinds of training paradigms that were possible for the PVM tracker.  Experiment 1 involved simultaneous long-running unsupervised training of the PVM network and supervised training of the tracker readout. Experiment 2 explored the question of whether the unsupervised portion of the PVM could be trained first, prior to a separate ``priming'' phase (as in existing trackers). In Experiment 2, the tracker readout layers were only trained during the final phase of learning.  Although we expected that the results might be slightly worse than with fully supervised training, we hypothesized that the bulk of the learning takes place in the unsupervised portion of the model --- and that final tracking performance should not be qualitatively different between the two experiments.

Experiment 3 tested the practical usability of the initial PVM trackers from Experiment 1 in an on-line environment.  PVM trackers from Experiment 1 with increasing amounts of training were embodied in a robot with a PID controller, and we quantified the time to reach a tracked target object.

\subsection{Experiment 1: Methods}

We trained a set of three separate PVM tracker instances, one on each training dataset (green ball, stop sign, and face), presenting it with sequences of labeled movies in a loop.  These movies were labeled as described in Section~\ref{sec:pvm_dataset}. The unsupervised PVM and the supervised tracker readout were trained simultaneously; the tracker readout heatmaps were trained to track the object.  When the target object was not present, all supervisory signals in the tracker readout layer were set to 0.5. Each model was trained on 10-20 min of repeated video (see Table~\ref{table_datasets}). 

The following learning rate schedule was used: Initially the training progressed layer-wise, where each subsequent layer was enabled 100,000 steps after the previous one.
For the first 100,000 steps of training of each layer the learning rate was set to 0.0002. 
After 100,000 steps the learning rate was lowered to 0.00005. 
After 700,000 steps all six layers of the hierarchy were enabled and the learning rate was set at 0.00005.
Lateral context was enabled at 700,000 steps and feedback connectivity was enabled at 900,000 steps. 
After 1.5M steps the learning rate in all layers was lowered again to a final value of 0.00001. 
The learning rate schedule that we selected appeared to speed up learning; however, having such a schedule does not appear to be critical for attaining the final performance levels we obtained. Instances of models run with other learning rate update schedules did not demonstrate any qualitative differences (data not shown). The learning rate schedule we report is not the result of an exhaustive search of meta-parameters space on these models, thus we do not make any claims about its optimality.

\label{sec:training-readout}

The PVM is outfitted for tracking by supplying an additional signal to be predicted by each association unit (see $M_{t+1}$ in Figure~\ref{fig:pvm_unit}). These additional signals were spatially arranged to align with the processed visual data and trained to activate whenever the object of interest was present in the area “seen” by each tile (this may be referred to as retinotopic or topographic projection). Thus, each layer of the PVM structure tries to predict a heatmap composed of these additional signals. 

It is important to note that during this simultaneous training of the unsupervised and supervised parts of the PVM tracker, the error from the readout layers was backpropagated  through the hidden layers of the PVM units.  Consequently, in this particular experiment, the tracking ``task'' may have influenced the internal representations developed throughout the PVM network.  (This is not the case for Experiment 2, see Section~\ref{sec:experiment-2}.)

\subsection{Experiment 1: Results}

\label{sec:experiment-1-results}

\begin{figure}[ht]
\centering
\includegraphics[width=0.9\textwidth]{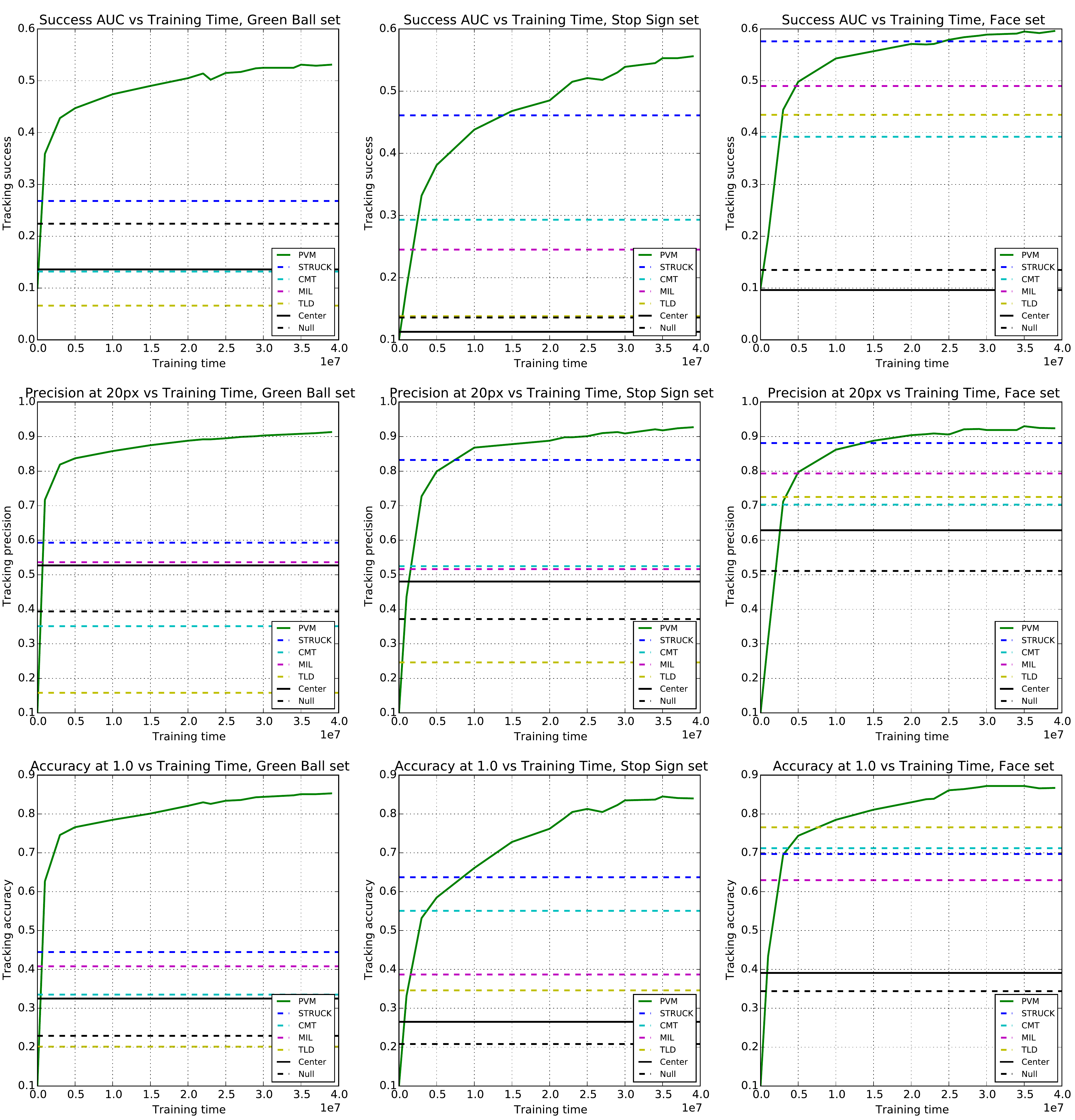}
\caption{\label{fig:Performance_sup} \textbf{Tracking Performance (Experiment 1)}. Object tracking performance of the PVM as a function of training time (in 10s of millions of steps), evaluated on the test set. Columns from left to right: green ball, stop sign and face datasets. Rows from top to bottom: area under the curve of Success plot, tracking Precision at 20 pixels and tracking Accuracy at 1.0. Performance of state of the art trackers displayed as horizontal lines, including STRUCK, CMT, MIL, TLD trackers. Center and null tracker are displayed to set the baseline. In all cases (all sets and all measures) after sufficient training (39M steps in this case) the PVM tracking performance exceeds the performance of state of the art trackers on the testing set. Note that the shown increases in performance are on the test set, which means that with increasing exposure to the training set, the model continues to extract relevant knowledge that generalizes well to the test set.}
\end{figure}

Instances of the PVM tracker trained after 1, 3, 5, 10, 15, 20, 23, 25, 27, 29, 30, 34, 35, 37, 39 million steps were evaluated on the test set with all learning frozen during each evaluation. The results of our evaluation are presented in the plot matrix in Figure \ref{fig:Performance_sup}. Note that the performance,  measured on the larger test set\footnote{Because the test set is larger than the training set, and because real world data is so highly variable, the test data is necessarily more diverse than the training set, making it a good probe of generalization.}, steadily increases. This indicates that the system is able to generalize well. Presented in the figure are also performance levels of other state of the art trackers as well as the control null and center trackers (which are a measure of bias in the data). In all three datasets and in three measures, with continued sufficient training the PVM is able to exceed the state of the art.  In the case of seemingly simple green ball dataset (left column) and the stop sign data set (middle column), the performance gain over the state of the art is substantial.

\begin{figure}[ht]
  \centering
  \begin{tabular}[c]{ccc}
    \begin{subfigure}[b]{0.32\textwidth}
      \includegraphics[width=\textwidth,trim={0cm 0.0cm 0cm 0.5cm},clip]{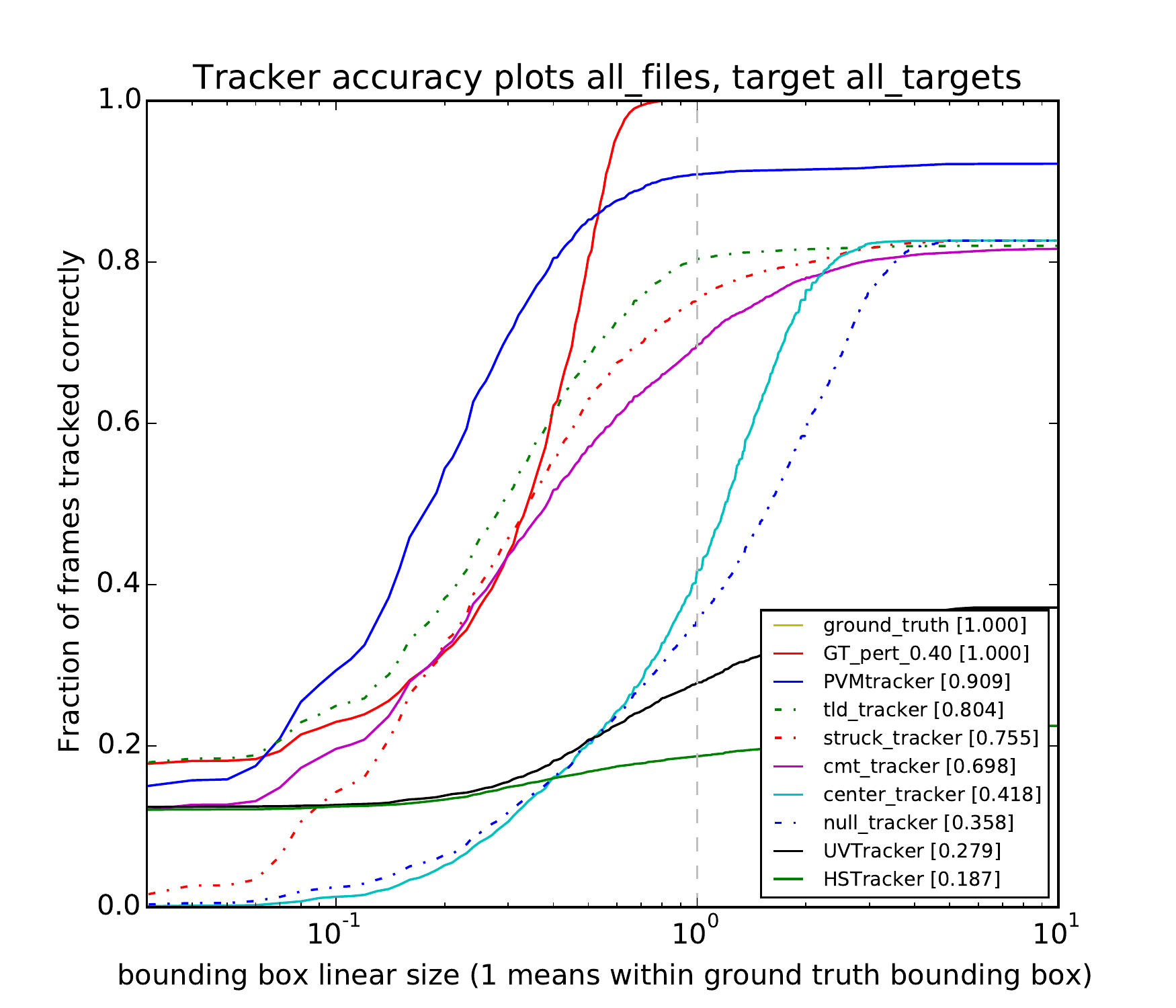}
      \caption{Accuracy}
      \label{fig:detailed_res_face:s}
    \end{subfigure}&
    \begin{subfigure}[b]{0.32\textwidth}
      \includegraphics[width=\textwidth,trim={0cm 0.0cm 0cm 0.5cm},clip]{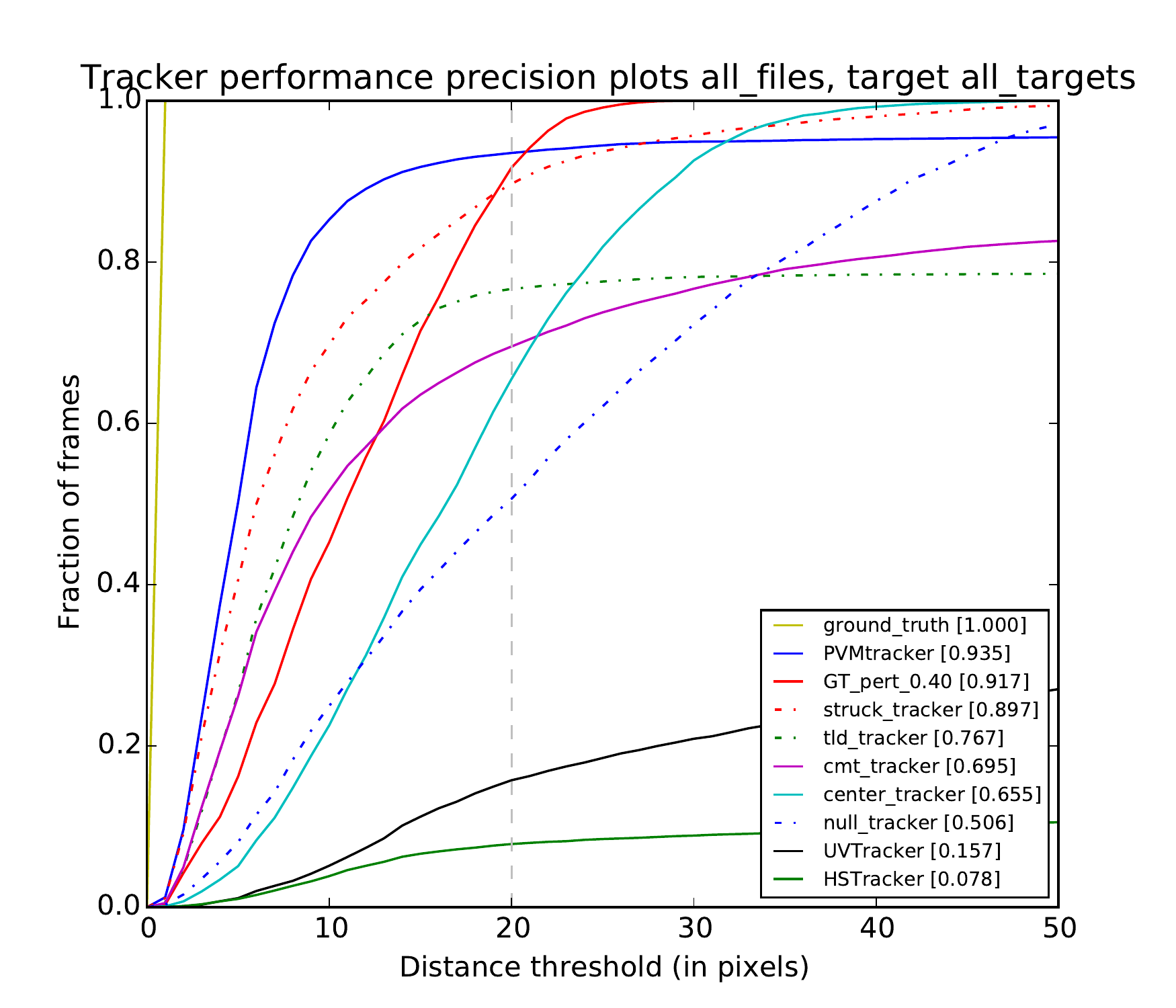}
      \caption{Precision}
      \label{fig:detailed_res_face:p}
    \end{subfigure}&
    \begin{subfigure}[b]{0.32\textwidth}
      \includegraphics[width=\textwidth,trim={0cm 0.0cm 0cm 0.5cm},clip]{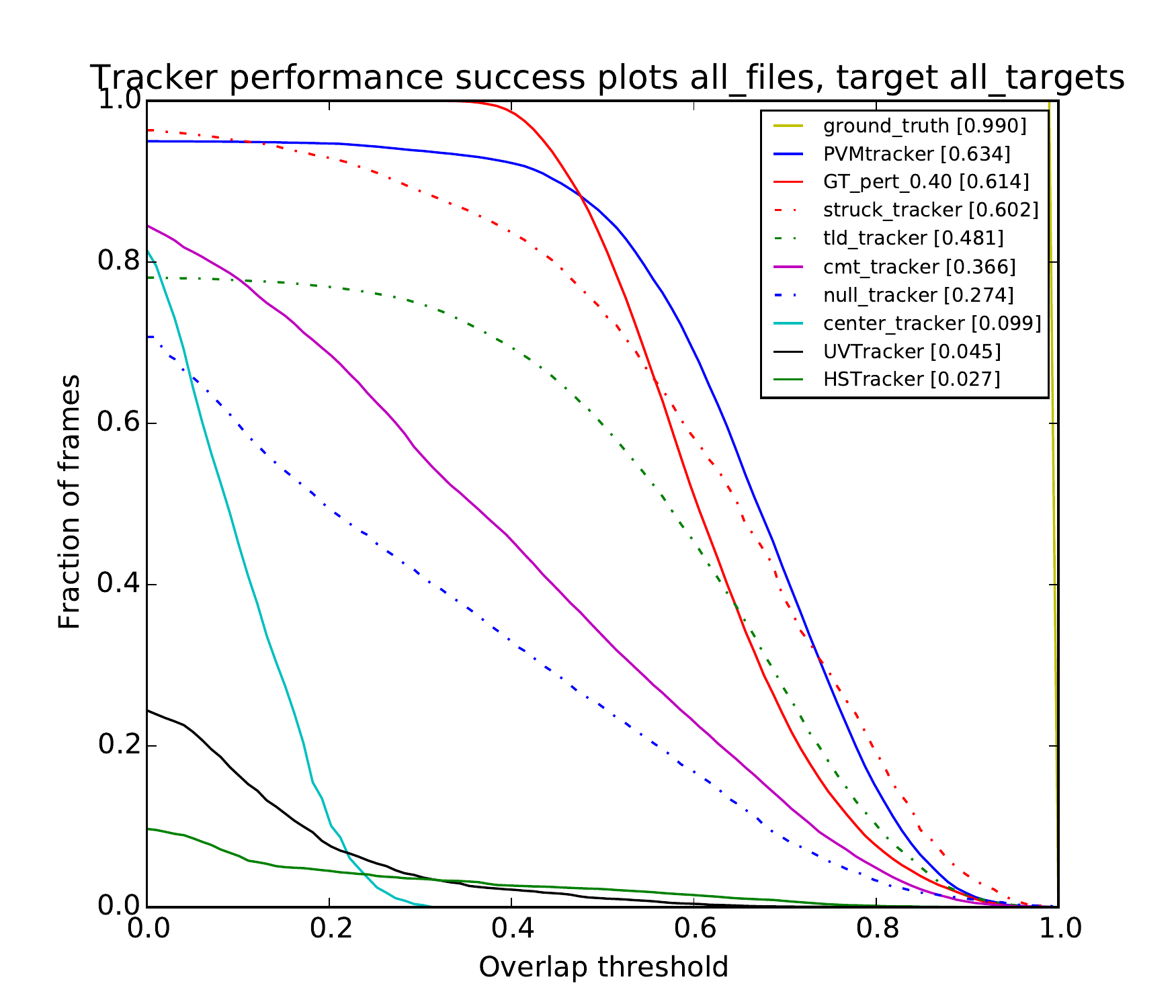}
      \caption{Success}
      \label{fig:detailed_res_face:f}
    \end{subfigure}\\
  \end{tabular}
  \caption{\textbf{Tracking results on the face dataset (Experiment 1)}. PVM tracker vs other state of the art trackers and ground truth perturbed by 40\% (to give the idea on the sensitivity of each measure). Ground truth (=1.0) overlaps with the axis. }\label{fig:detailed_res_face}
\end{figure}

\begin{figure}[ht]
  \centering
  \begin{tabular}[c]{ccc}
    \begin{subfigure}[b]{0.32\textwidth}
      \includegraphics[width=\textwidth,trim={0cm 0.0cm 0cm 0.5cm},clip]{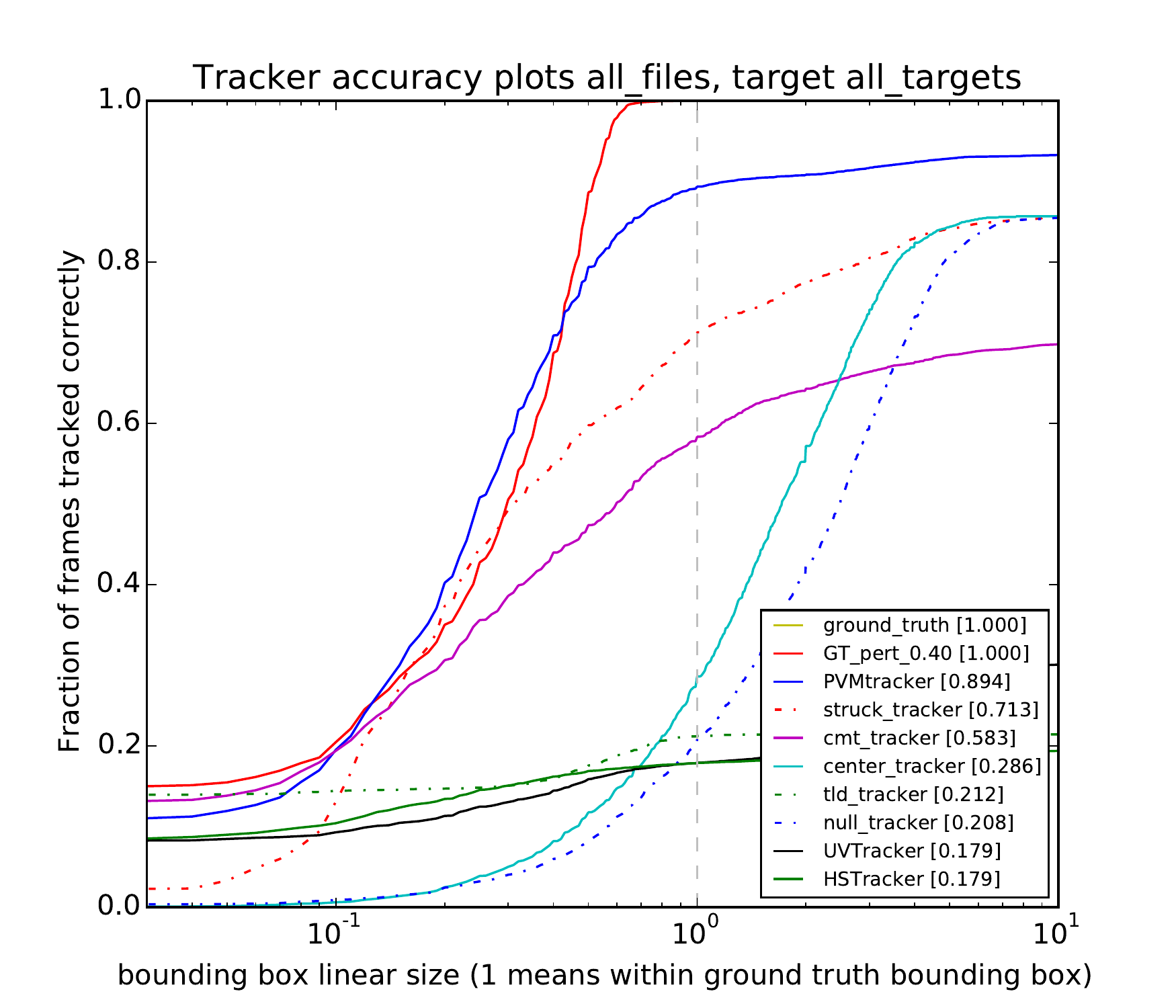}
      \caption{Accuracy}
      \label{fig:detailed_res_stop:a}
    \end{subfigure}&
    \begin{subfigure}[b]{0.32\textwidth}
      \includegraphics[width=\textwidth,trim={0cm 0.0cm 0cm 0.5cm},clip]{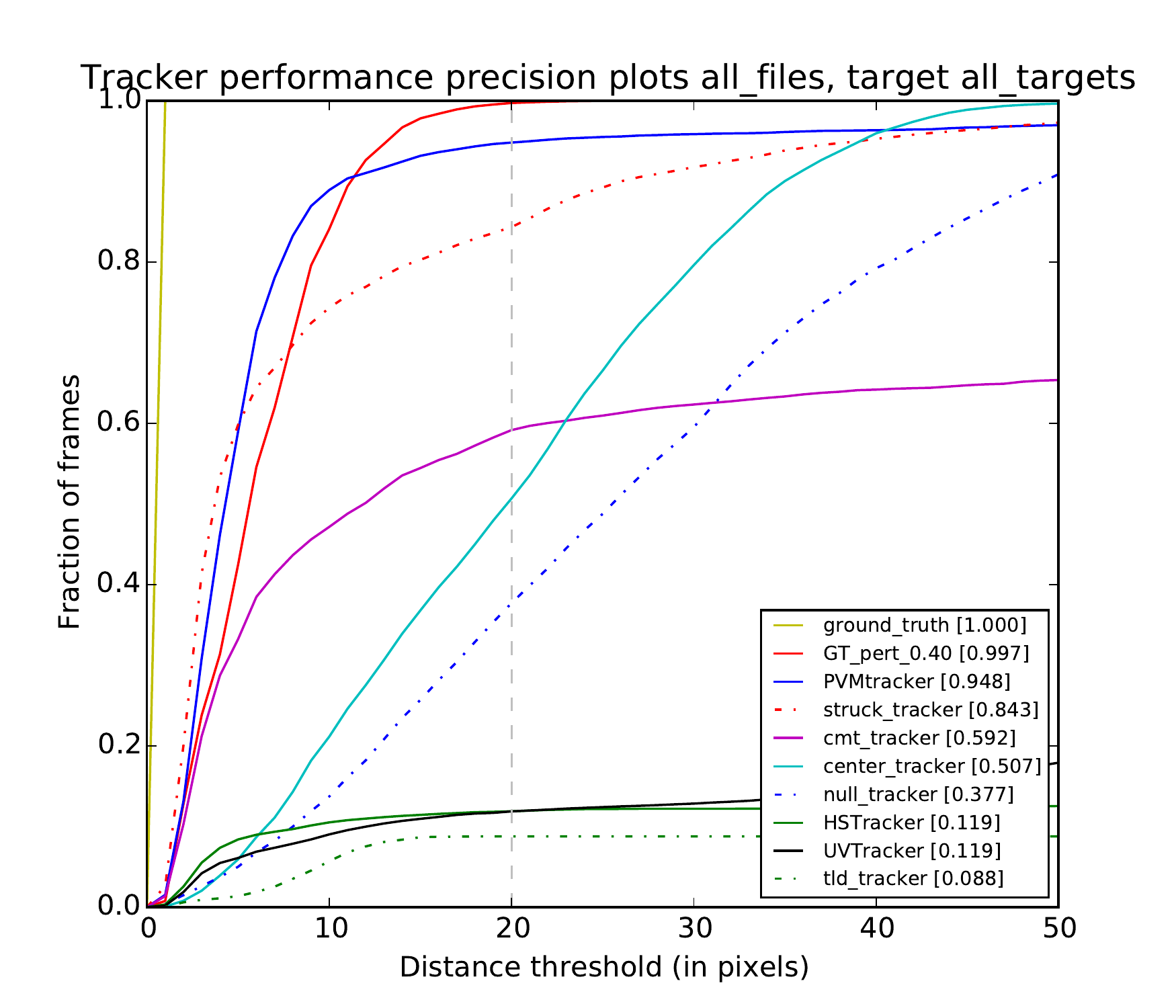}
      \caption{Precision}
      \label{fig:detailed_res_stop:p}
    \end{subfigure}&
    \begin{subfigure}[b]{0.32\textwidth}
      \includegraphics[width=\textwidth,trim={0cm 0.0cm 0cm 0.5cm},clip]{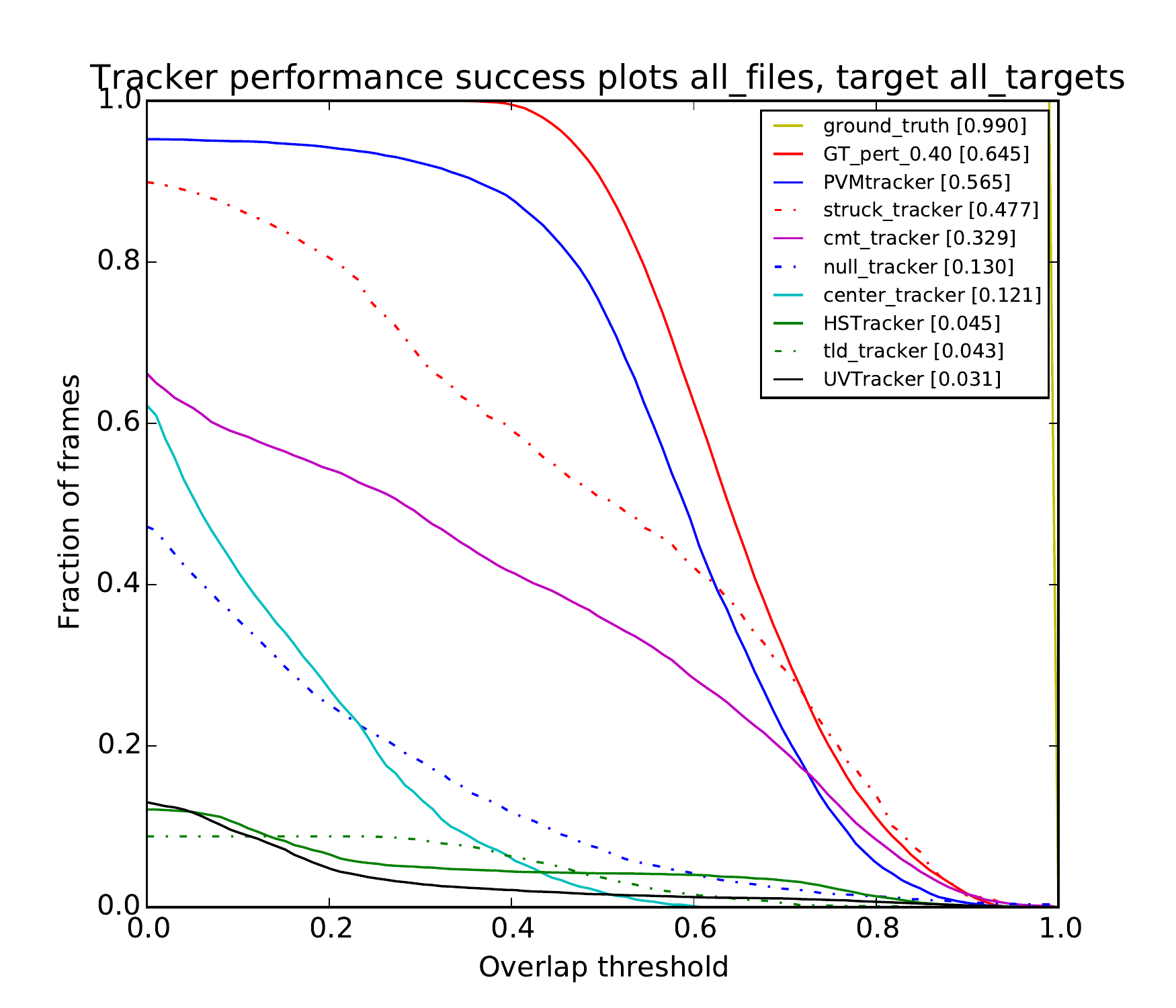}
      \caption{Success}
      \label{fig:detailed_res_stop:s}
    \end{subfigure}\\
  \end{tabular}
  \caption{\textbf{Tracking results on the stop sign dataset (Experiment 1).} PVM tracker vs other state of the art trackers and ground truth perturbed by 40\% (to give the idea on the sensitivity of each measure).  Ground truth (=1.0) overlaps with the axis.}\label{fig:detailed_res_stop}
\end{figure}

\begin{figure}[ht]
  \centering
  \begin{tabular}[c]{ccc}
    \begin{subfigure}[b]{0.32\textwidth}
      \includegraphics[width=\textwidth,trim={0cm 0.0cm 0cm 0.5cm},clip]{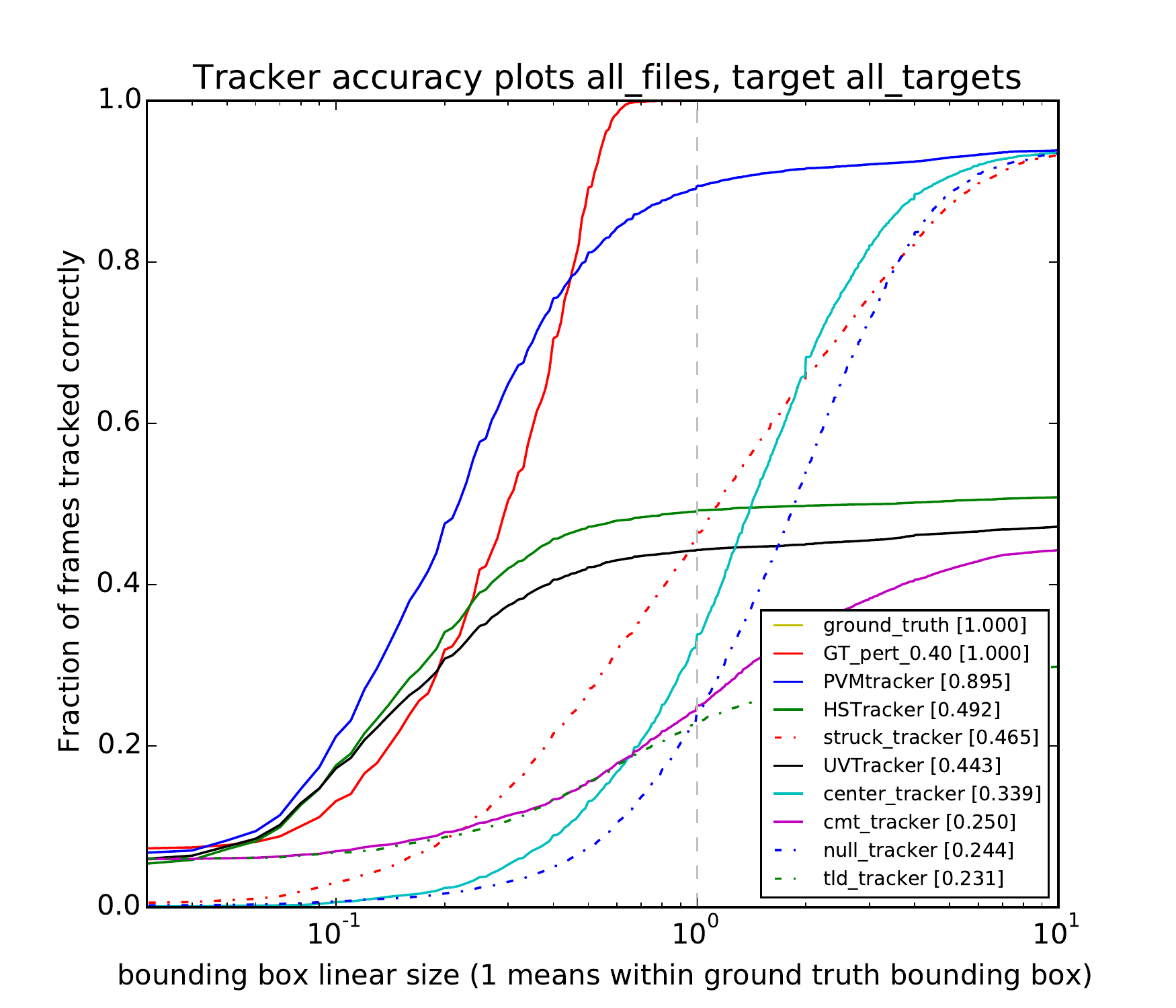}
      \caption{Accuracy}
      \label{fig:detailed_res_ball:a}
    \end{subfigure}&
    \begin{subfigure}[b]{0.32\textwidth}
      \includegraphics[width=\textwidth,trim={0cm 0.0cm 0cm 0.5cm},clip]{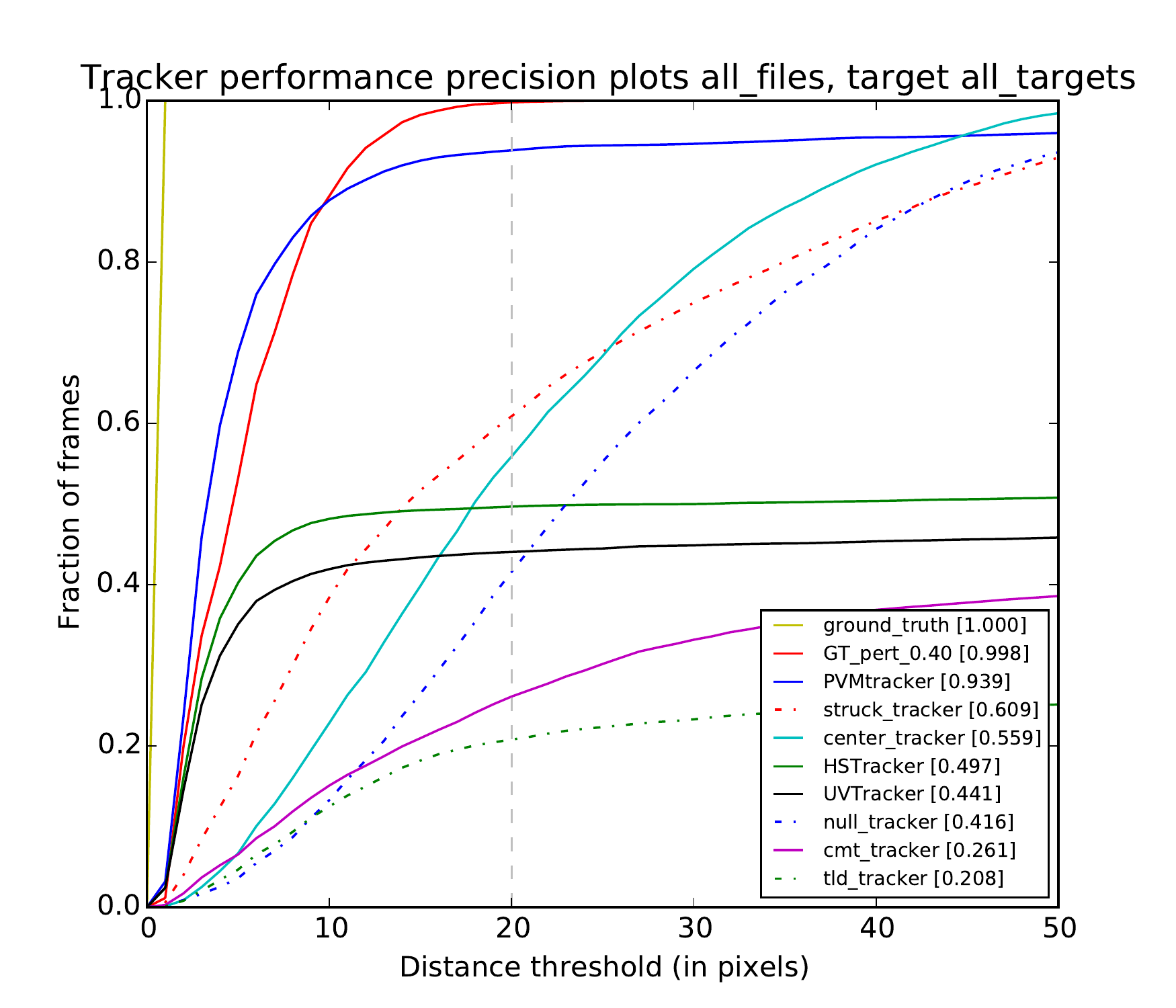}
      \caption{Precision}
      \label{fig:detailed_res_ball:p}
    \end{subfigure}&
    \begin{subfigure}[b]{0.32\textwidth}
      \includegraphics[width=\textwidth,trim={0cm 0.0cm 0cm 0.5cm},clip]{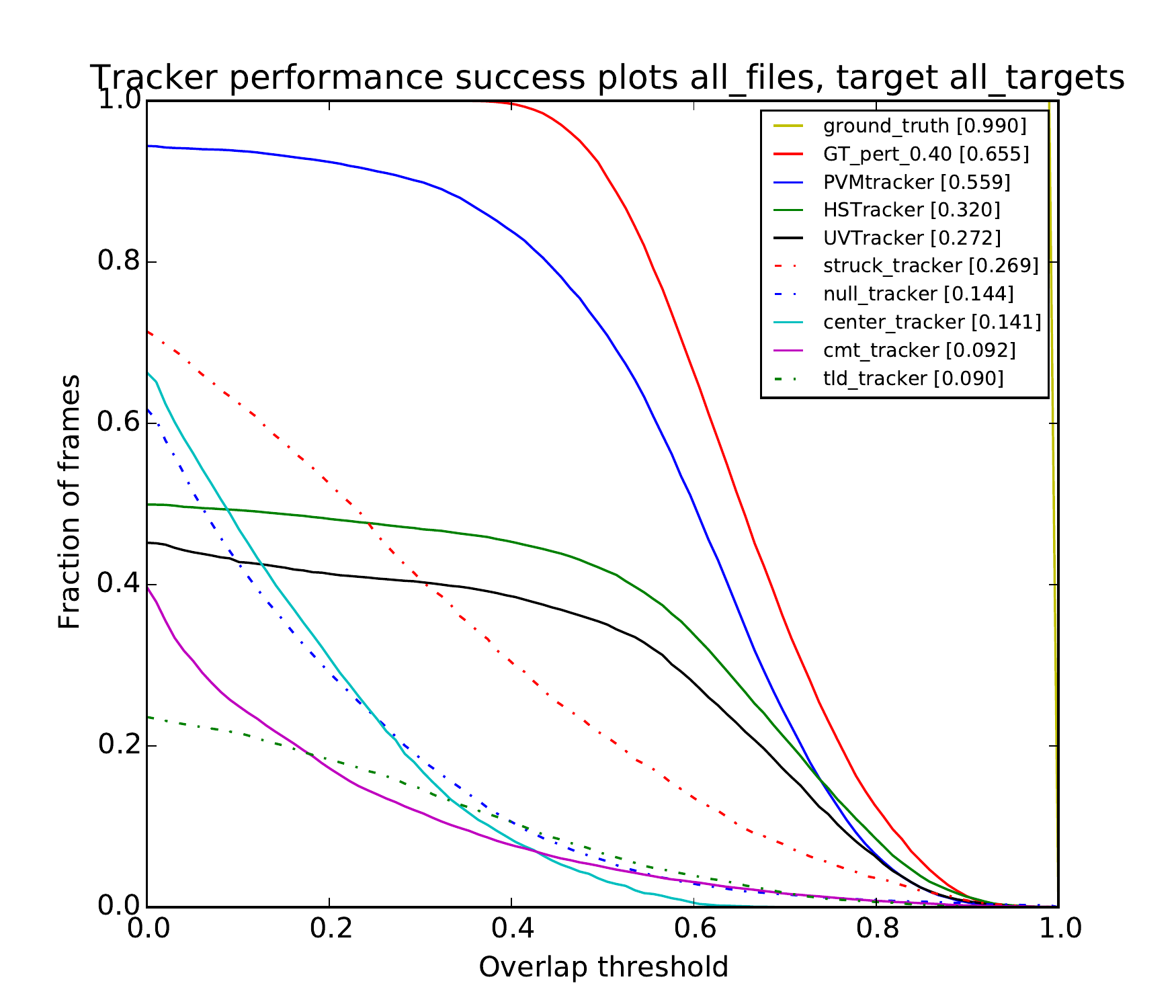}
      \caption{Success}
      \label{fig:detailed_res_ball:s}
    \end{subfigure}\\
  \end{tabular}
  \caption{\textbf{Tracking results on the green ball sign dataset (Experiment 1)}. PVM tracker vs other state of the art trackers and ground truth perturbed by 40\% (to give the idea on the sensitivity of each measure).  Ground truth (=1.0) overlaps with the axis.}\label{fig:detailed_res_ball}
\end{figure}

\begin{figure}[ht]
  \centering
  \begin{tabular}[c]{ccc}
    \begin{subfigure}[b]{0.32\textwidth}
      \includegraphics[width=\textwidth, page=9]{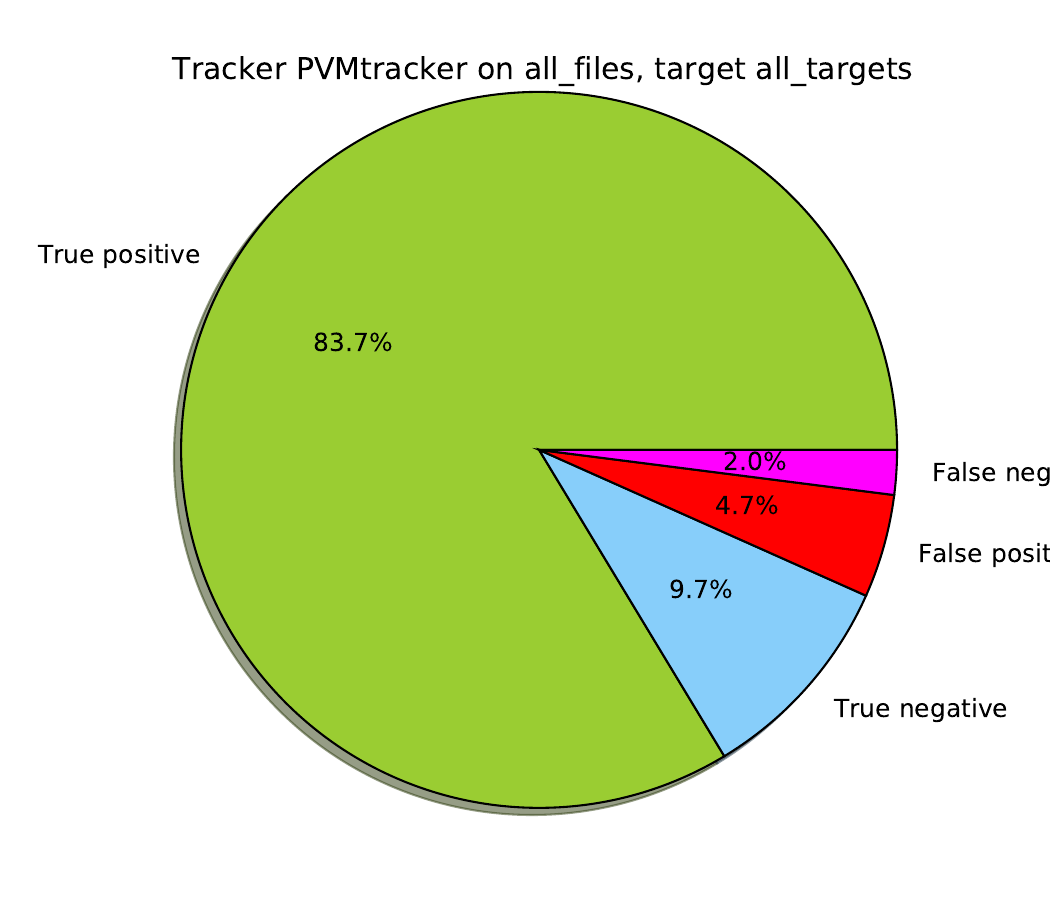}
      \caption{PVM: Stop sign set}
      \label{fig:pie_charts:pvm_s}
    \end{subfigure}&
    \begin{subfigure}[b]{0.32\textwidth}
      \includegraphics[width=\textwidth, page=9]{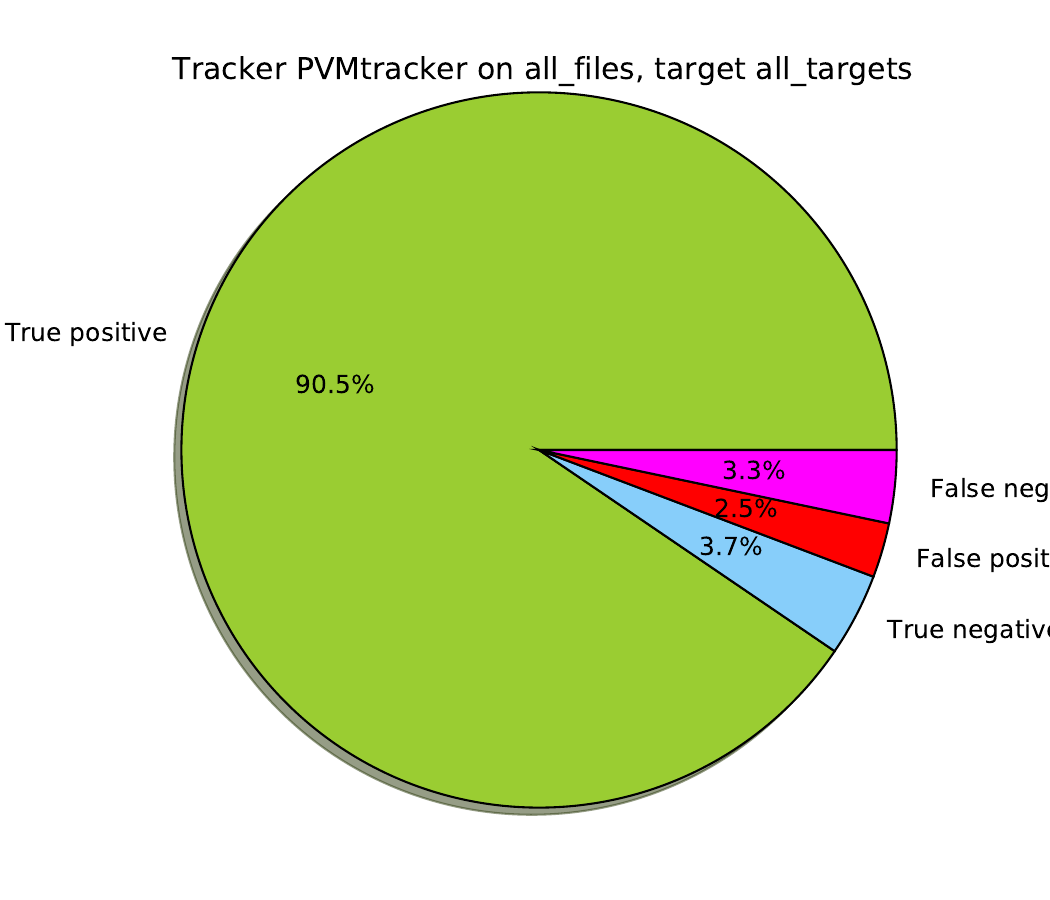}
      \caption{PVM: Green ball set}
      \label{fig:pie_charts_pvm:g}
    \end{subfigure}&
    \begin{subfigure}[b]{0.32\textwidth}
      \includegraphics[width=\textwidth, page=9]{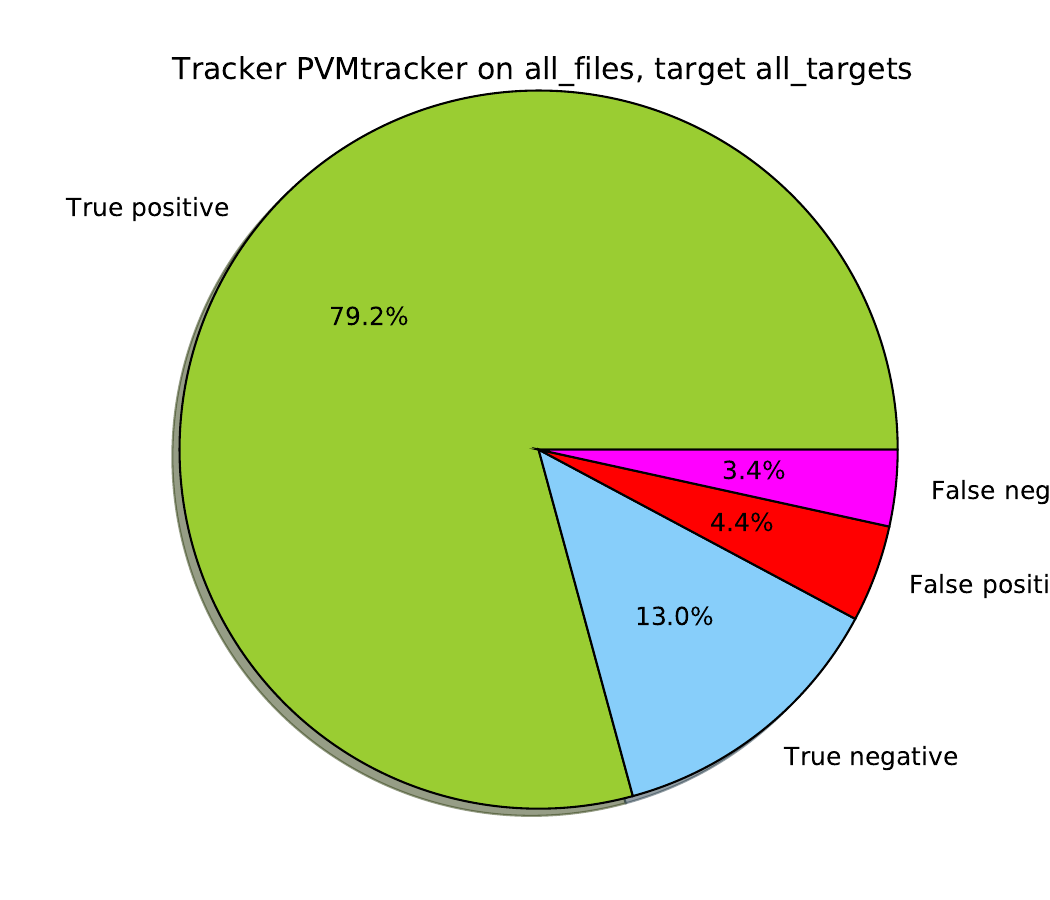}
      \caption{PVM: Face set}
      \label{fig:pie_charts_pvm_f}
    \end{subfigure}\\
    \begin{subfigure}[b]{0.32\textwidth}
      \includegraphics[width=\textwidth, page=8]{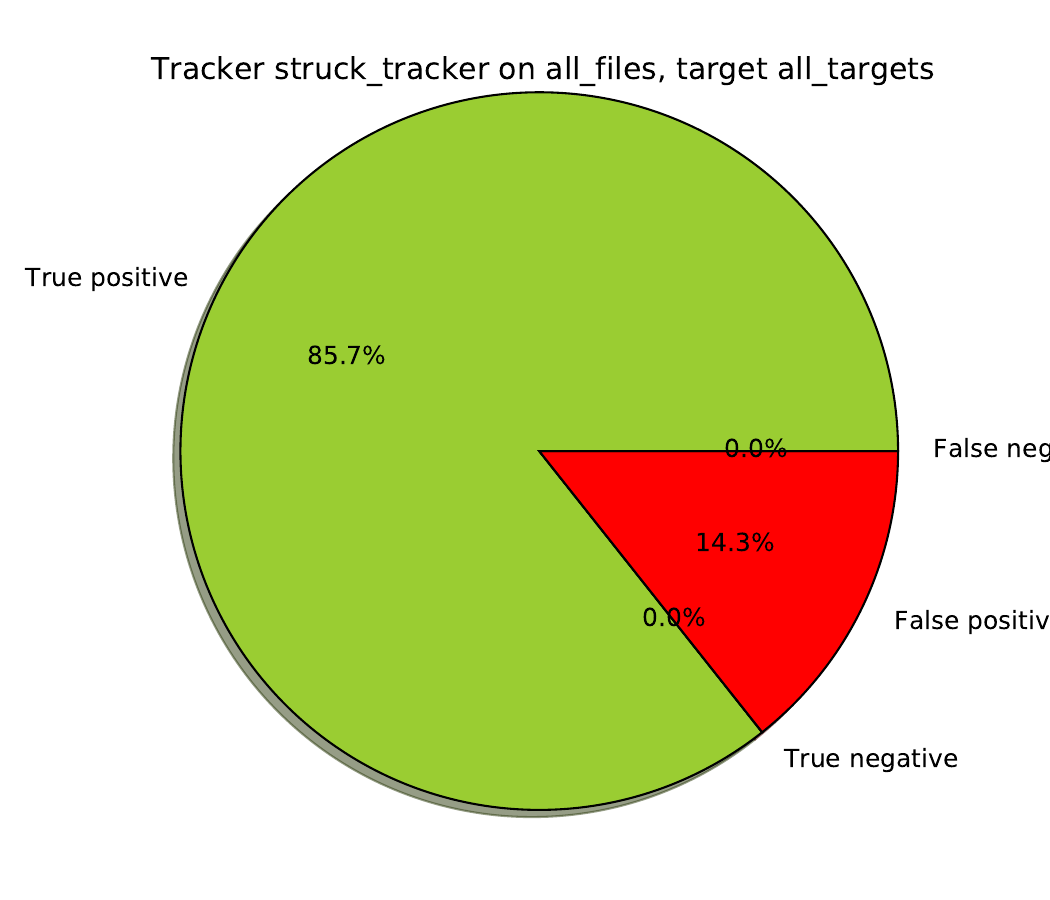}
      \caption{STRUCK: Stop sign set}
      \label{fig:pie_charts_struck_s}
    \end{subfigure}&
    \begin{subfigure}[b]{0.32\textwidth}
      \includegraphics[width=\textwidth, page=8]{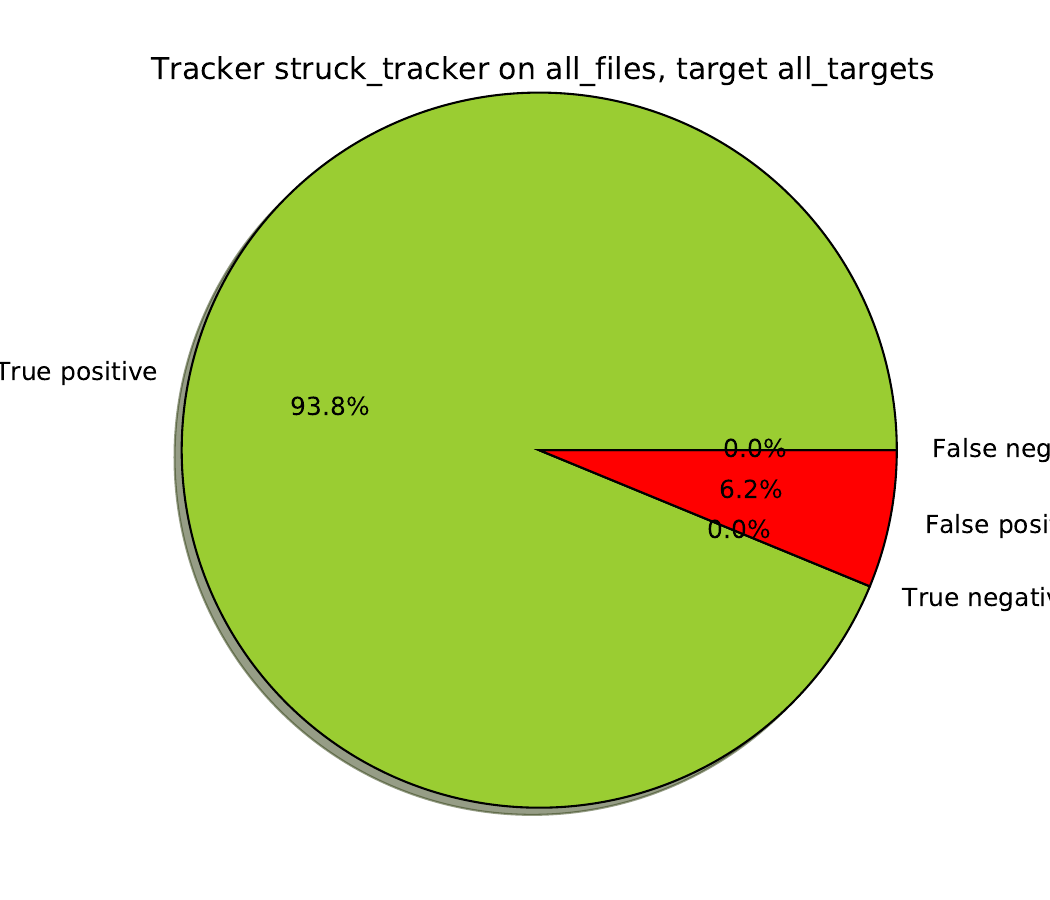}
      \caption{STRUCK: Green ball set}
      \label{fig:pie_charts_struck_b}
    \end{subfigure}&
    \begin{subfigure}[b]{0.32\textwidth}
      \includegraphics[width=\textwidth, page=8]{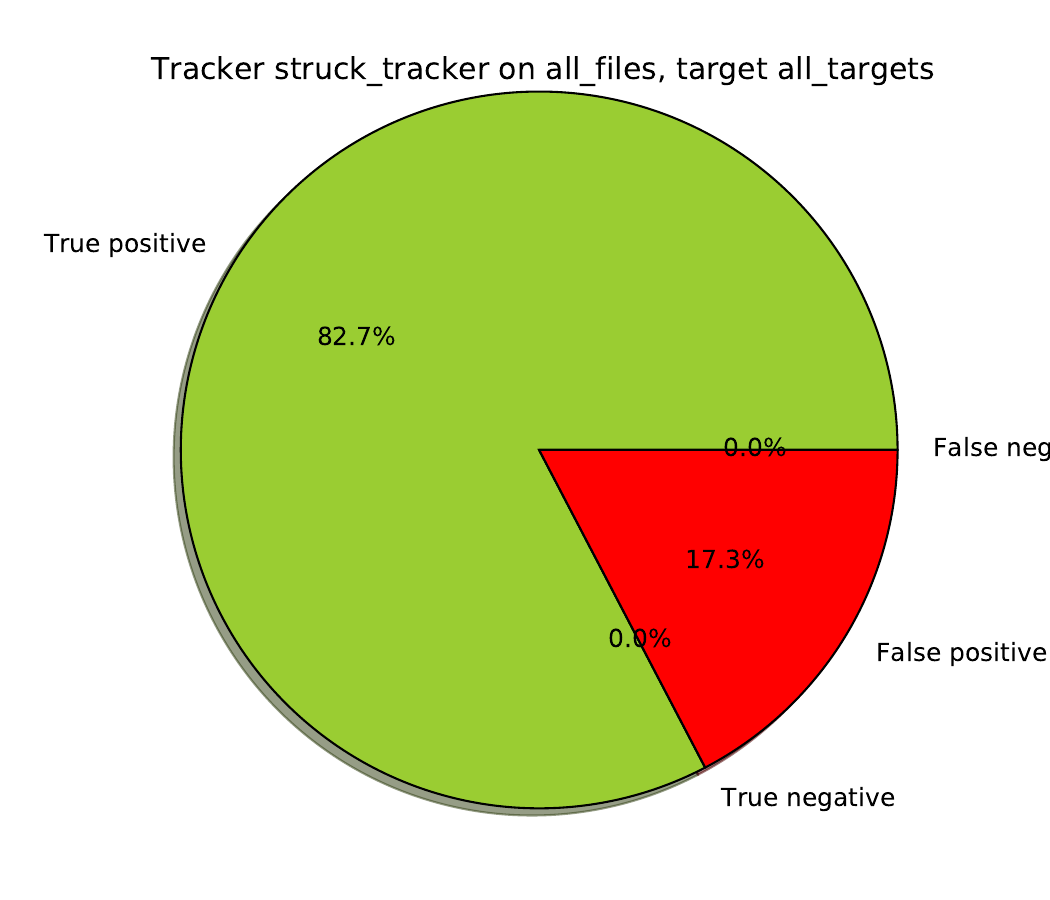}
      \caption{STRUCK: Face set}
      \label{fig:pie_charts_struck_f}
    \end{subfigure}\\
    \begin{subfigure}[b]{0.32\textwidth}
      \includegraphics[width=\textwidth, page=2]{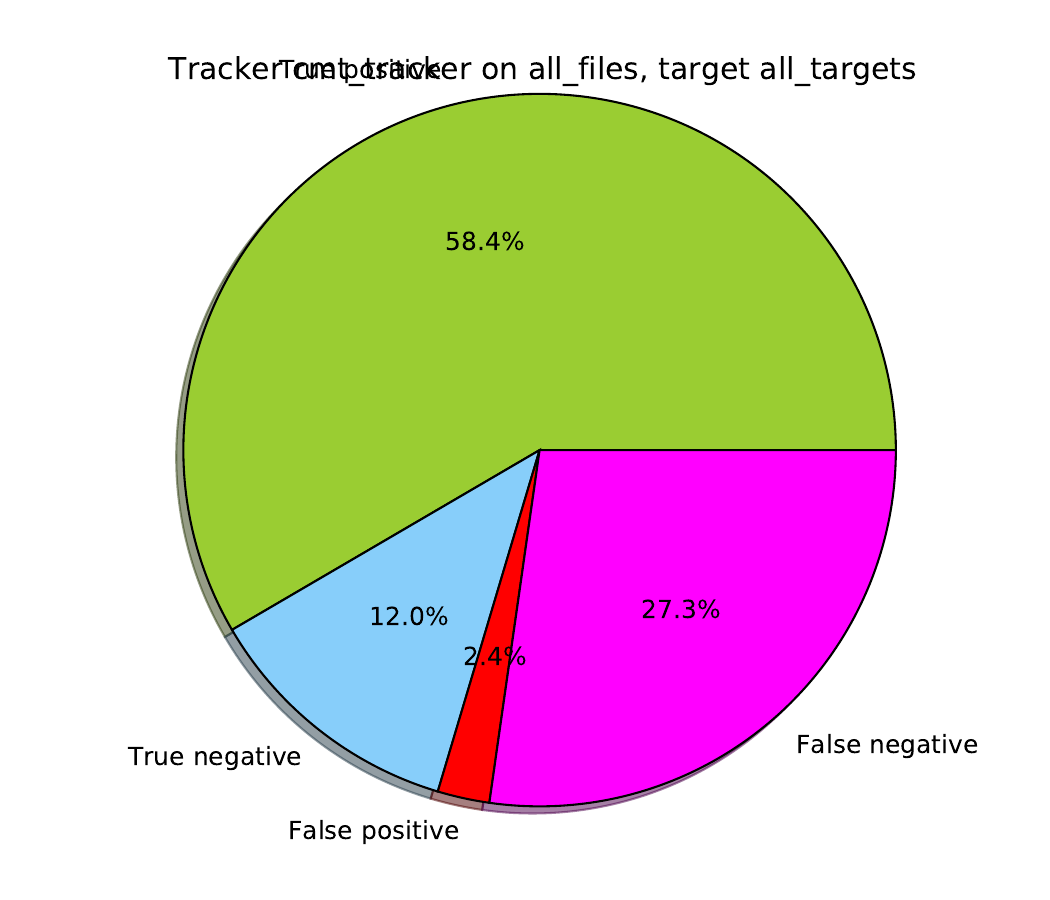}
      \caption{CMT: Stop sign set}
      \label{fig:pie_charts_cmt_s}
    \end{subfigure}&
    \begin{subfigure}[b]{0.32\textwidth}
      \includegraphics[width=\textwidth, page=2]{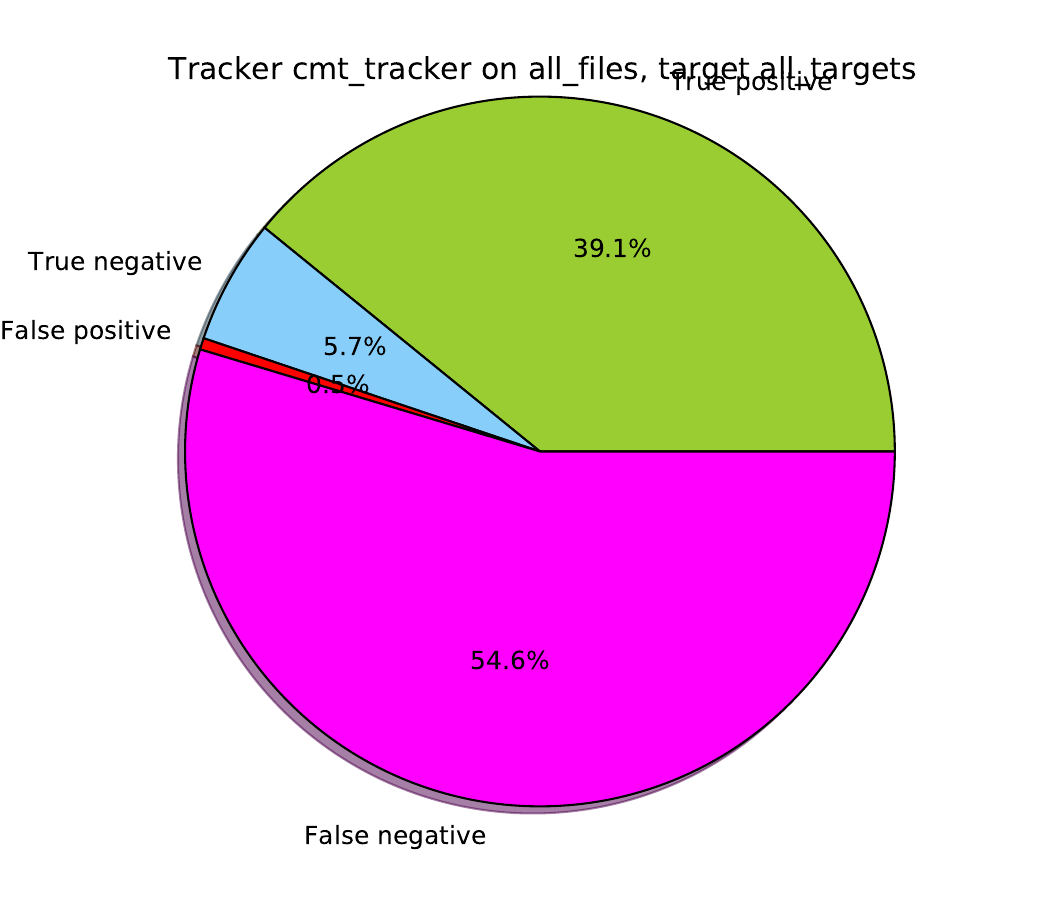}
      \caption{CMT: Green ball set}
      \label{fig:pie_charts_cmt_g}
    \end{subfigure}&
    \begin{subfigure}[b]{0.32\textwidth}
      \includegraphics[width=\textwidth, page=2]{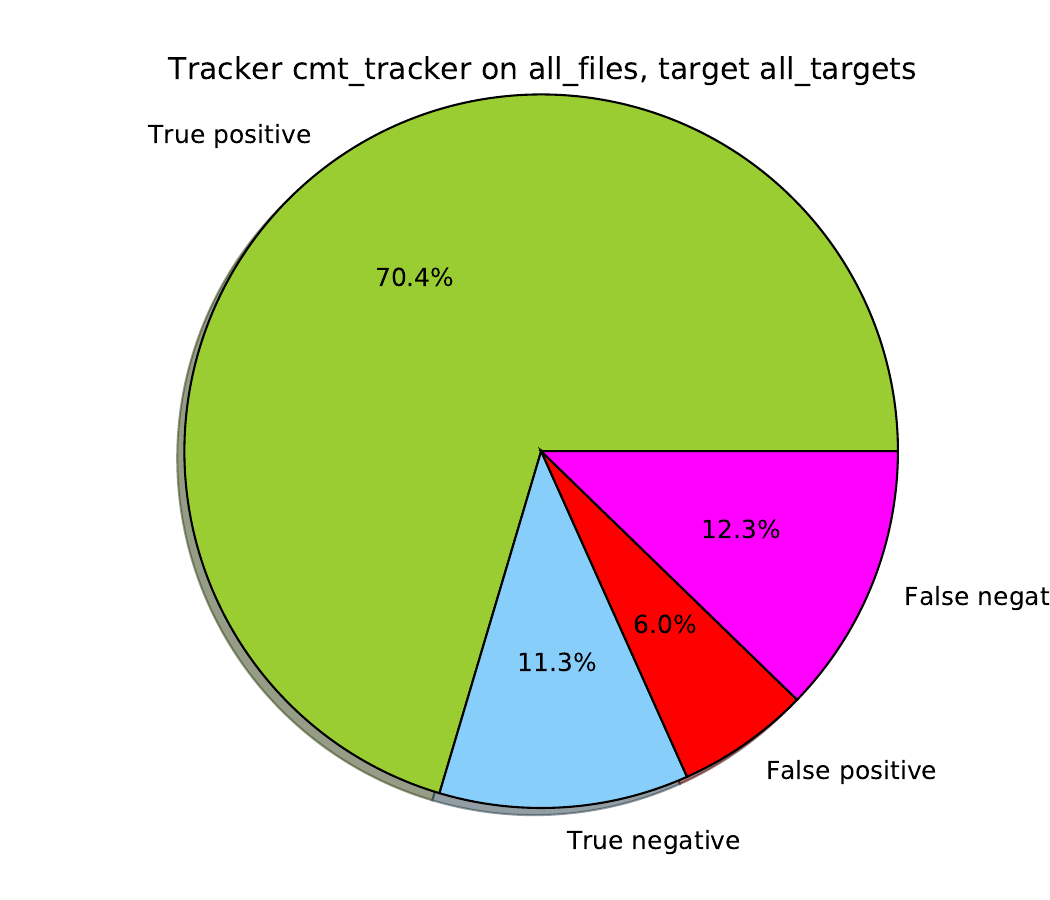}
      \caption{CMT: Face set}
      \label{fig:pie_charts_cmt_f}
    \end{subfigure}\\

\end{tabular}
  \caption{\textbf{Target Presence Detection Performance (Experiment 1)}. Classification performance on target presence for PVM (upper row), STRUCK (middle row) and CMT (lower row) trackers. Ideal tracker should only contain the green (true positive) and the blue (true negative) pies. STRUCK does not detect negatives at all and always reports the target, hence a sizable false positive portion in the pie chart. CMT on the other hand often does not recognize the target, hence substantial false negative part. }\label{fig:pie_charts}
\end{figure}

\begin{figure}[ht]
  \centering
  \begin{tabular}[c]{c}
    \begin{subfigure}[b]{\textwidth}
    \centering
      \includegraphics[width=0.15\textwidth]{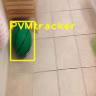}
      \includegraphics[width=0.15\textwidth]{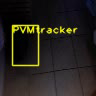}
      \includegraphics[width=0.15\textwidth]{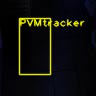}
      \includegraphics[width=0.15\textwidth]{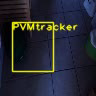}
      \includegraphics[width=0.15\textwidth]{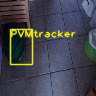}
      \caption{Switching off light, green ball set} \label{fig:gb_examples}    
    \end{subfigure}
\\
    \begin{subfigure}[b]{\textwidth}
    \centering
      \includegraphics[width=0.15\textwidth]{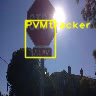}
      \includegraphics[width=0.15\textwidth]{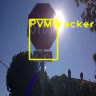}
      \includegraphics[width=0.15\textwidth]{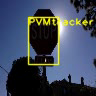}
      \includegraphics[width=0.15\textwidth]{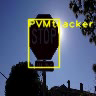}
      \includegraphics[width=0.15\textwidth]{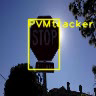}
      \caption{Lens flare, exposure compensation, stop sign set} \label{fig:ss_examples}     
    \end{subfigure}
\\
    \begin{subfigure}[b]{\textwidth}
    \centering
      \includegraphics[width=0.15\textwidth]{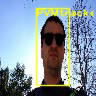}
      \includegraphics[width=0.15\textwidth]{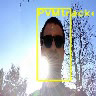}
      \includegraphics[width=0.15\textwidth]{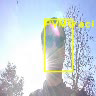}
      \includegraphics[width=0.15\textwidth]{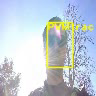}
      \includegraphics[width=0.15\textwidth]{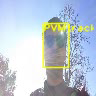}
      \caption{Lens flare, face set} \label{fig:ff_examples}    
    \end{subfigure}
\\
  \end{tabular}
  \caption{\textbf{Examples of Tracking through Challenging Lighting Transitions (Experiment 1)}. 
  Every other frame of the sequence shown. Examples taken from the test set with PVM instances trained for 96M steps on each object respectively. PVM handles these massive appearance transitions very well.}\label{fig:examples}
\end{figure}

More detailed plots showing the performance of all trackers after 95M steps of supervised training are presented in Figures \ref{fig:detailed_res_face}, \ref{fig:detailed_res_stop}, and \ref{fig:detailed_res_ball}. 
As noted in Section~\ref{sec:tracker_metrics}, each of the reported measures is a function of a parameter.  The plots show the values of these measures as the parameters $\theta$, $\rho$, and $\phi$ are varied. To get a better idea of the extent to which these results compare to human-like performance, an additional artificial tracker was added to these plots marked as GT\_per\_0.4. This tracker is the ground truth perturbed by $40\%$ both in position and size (i.e., the position and size are randomly and uniformly perturbed within $40\%$ of the given dimension of the original box). We see that PVM tracker performance closely approaches the perturbed ground truth, indicating a very good tracking performance.

Figure \ref{fig:pie_charts} shows another important aspect of visual tracking performance, specifically how well the tracker correctly reports the presence or absence of the target object. This important characteristic is not well captured by either Success, Precision or Accuracy measures. We specifically created our datasets to contain some fraction (typically $10-20\%$) of frames without the target. These were included to measure false positives, which can be a very important mode of failure, particularly for embodied robotic devices. We can see in Figure \ref{fig:pie_charts} that PVM obtains good results with this measure as well, covering the great majority of true positives and almost $2/3$ of true negatives. 

\FloatBarrier

\subsection{Experiment 1: Discussion}

Overall the PVM tracking results are very good (see Figure~\ref{fig:examples} for a few examples of particularly challenging transitions).
The obtained results could likely be further improved through various ad hoc methods like tweaking the parameters of the final classifier (bounding box generator) e.g., via ROC curve analysis and tinkering with the details of the algorithm.  However, we decided to leave that part of the tracker unchanged to avoid the temptation of injecting priors to artificially inflate our results. 

We noted features of the other trackers that impede their performance. The STRUCK tracker does not report true negatives at all --- it always tracks whatever is most similar to the original object. In contrast, CMT does make a decision and reports the majority of true negatives but at the expense of a significant fraction of false negatives (does not report the object when it in fact is there). 

\subsection{Experiment 2: Method}
\label{sec:experiment-2}

The second methodology involved pre-training the unsupervised PVM portion of the PVM tracker network on a single large, unlabeled video data set.  The large unlabeled training set included videos from all 3 datasets (green ball, stop sign, and human face) as well as additional general unlabeled video taken within the Brain Corporation offices.  The pre-trained network was trained according to the same learning rate schedule as the entire PVM tracker in Experiment 1. Instances of the pre-trained network were then ``primed'' or ``cued'' on a target object to track via supervised learning on a small, labeled, video dataset\footnote{This priming data is actually the same data as used to train the models in Experiment 1, just shown for a fraction of the training time}. Importantly, during this priming phase, learning only occurred in weights from the (frozen) unsupervised PVM network to the supervised readout layers trained based on the labeled bounding boxes. Three instances of each network were tested, with readout supervised learning rates of 0.001, 0.0005 and 0.0002.  Tracking performance was then evaluated on the network.  We emphasize that the lengthy, unsupervised training phase is performed only once: the various short, supervised training phases that specify the target object are all based on the same instance of the unsupervised network.

\subsection{Experiment 2: Results}
\begin{figure}[ht!]
\centering
\includegraphics[width=0.9\textwidth]{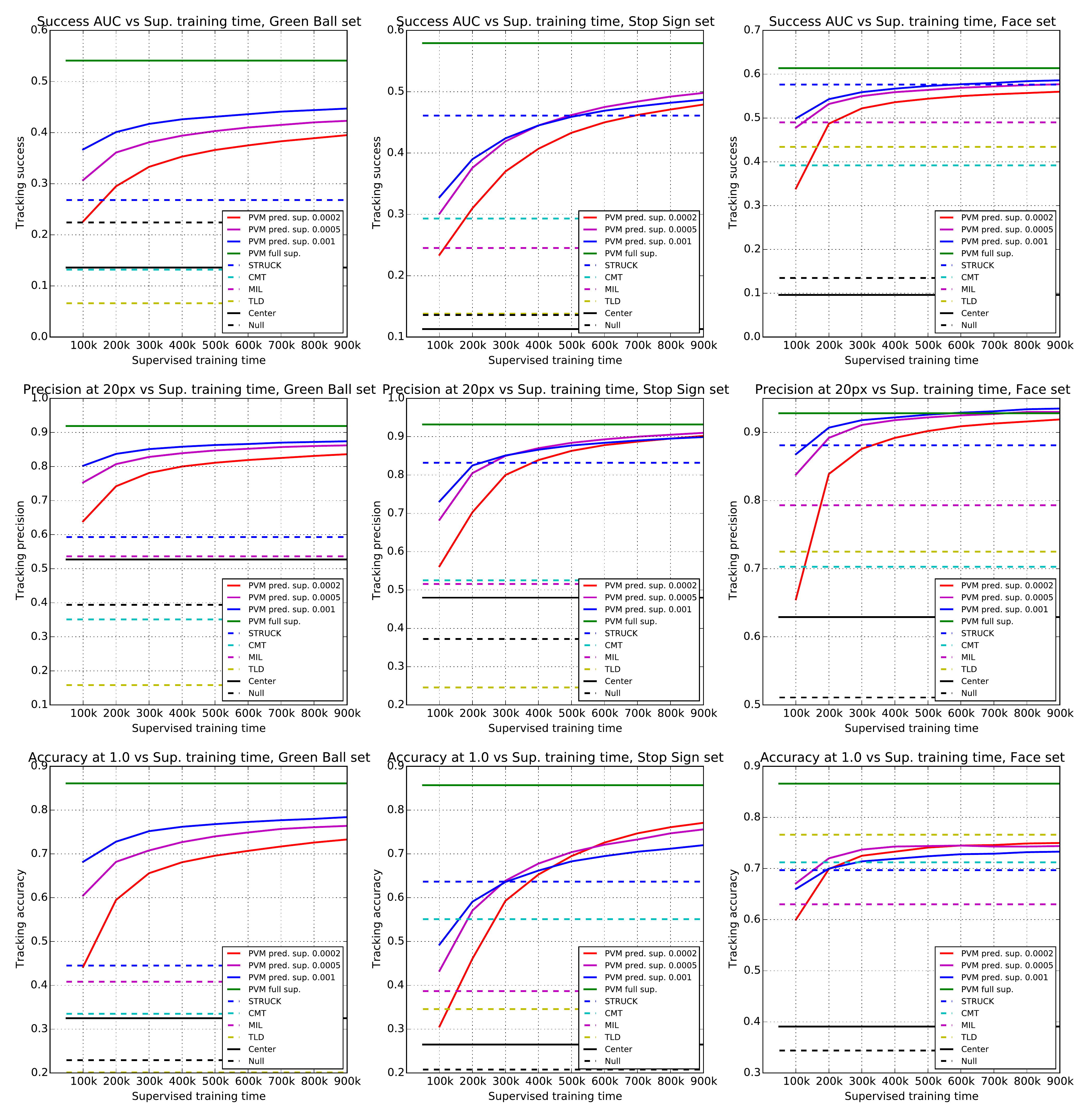}
\caption{\label{fig:Performance_unsup_sup}\textbf{Tracking Performance of the PVM tracker as a Function of Supervised Training Time (Experiment 2).} The model was first trained in an unsupervised manner for 50M steps. Then the predictive part of the network was frozen and the readout part was trained in supervised mode for 1M steps. The green line indicates the performance of the PVM tracker trained for 50M steps according to the methodology of Experiment 1. 
}
\end{figure}

Figure \ref{fig:Performance_unsup_sup} plots the tracking performance as a function of the shorter supervised training time for models that had previously completed a long unsupervised learning phase (50M time-steps). Results for the three supervised learning rates are shown. Recall that learning is only performed along the weights leading to the readout classifier; the predictive unsupervised part of the model is frozen. As expected, the results are slightly worse than the fully supervised PVM tracker (shown as green), but in many cases still better than the state-of-the-art trackers.

\subsection{Experiment 2: Discussion}

The results suggest that the unsupervised PVM portion of the PVM tracker is the locus of the bulk of the learning necessary to obtain good tracking.  The results also suggest that the PVM network is extracting something about the physical dynamics of the scenes and objects within them. We hypothesize that this physical knowledge is something that must be present and universal across the several training datasets.  If this hypothesis is true, the system will not be susceptible to catastrophic interference during subsequent learning, as long as the new data obeys the same physical dynamics. Further research is required to determine the nature of this learned information.

\subsection{Experiment 3: Method}
\label{section:virtual}

To study the usability of the PVM tracker in a real-time environment, and to demonstrate its robustness across conditions, we integrated it in a virtual-world robotic target approach task. The virtual world we use is an indoor environment with realistic wall, floor, and ceiling textures, realistic lighting and shadows, and a 3D physical model of a differential-drive robot. The environment is configured and simulated with Unreal\texttrademark\  Engine 4.9.  The dimensions of the virtual room were $400\times 400$cm, and the room had a wood texture floor with windows letting in simulated sunlight.  The virtual room also had internal spot lighting mounted on the ceiling. The target object, a virtual green basketball, measured approximately 40cm in diameter. 

Instances of the PVM tracker from Experiment 1 trained for 0.1, 0.2, 0.4, 0.6, 1, 2, 4, 8, 16, 32, 64 and 95 million time-steps on the green basketball dataset were used during the test.  Each PVM tracker instance was run on a local workstation with multiple processor cores comparable to the CPUs described in the Methods (Section~\ref{sec:execution-time-scalability}).  Because the model was no longer trained, but just executed, it was able to run in real time.

\begin{figure}[ht]
\centering
\includegraphics[width=0.9\textwidth,clip]{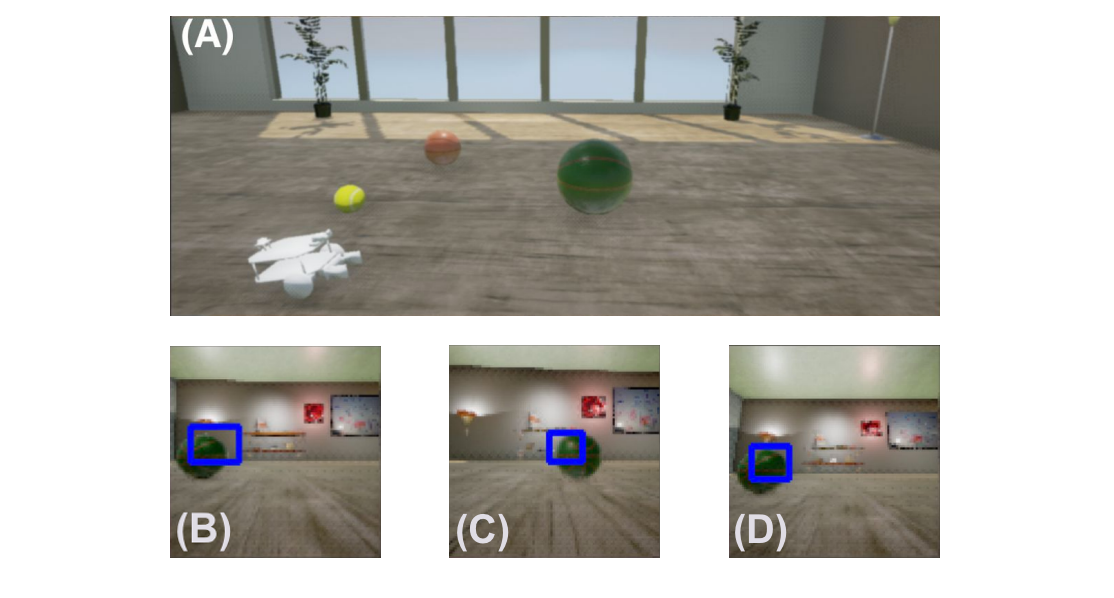}
  \caption{\label{fig:vw_setup01}\textbf{PVM Tracker on a Virtual Robot}. An instance of the PVM tracker supervised to track a green basketball was embodied in a virtual robot in a photorealistic simulated world using Unreal Engine 4. (A) Lateral view of robot with a green basketball and some distractor objects. A specular reflection can be seen on the green ball from the overhead lighting. Realistic lighting and shadows also come from the window. (B-D) Bounding boxes provided by the tracker as the robot approaches the green ball.}
\end{figure}

The robot model was created in Blender 2.74 and measured 30.5cm.  The robot was configured with a PID controller and steering that was driven by the center of the bounding box returned by the PVM tracker. Figure \ref{fig:vw_setup01} shows the virtual environment, robot, and bounding box. On each trial, the robot was placed at a fixed location in the room and the target object (green ball) was placed at random locations with distance from target to robot between 50 and 100 cm in an 80-degree range in view of the robot. We measured performance of the real-world approach behavior as a function of training time of the PVM tracker. Performance is defined as the percentage of trials where the robot successfully bumped into the target within a time limit of 20 seconds. Trials exceeding 20 seconds were timed-out and marked as failures.

\subsection{Experiment 3: Results}
\begin{figure}[ht]\centering\includegraphics[width=0.5\textwidth,clip]{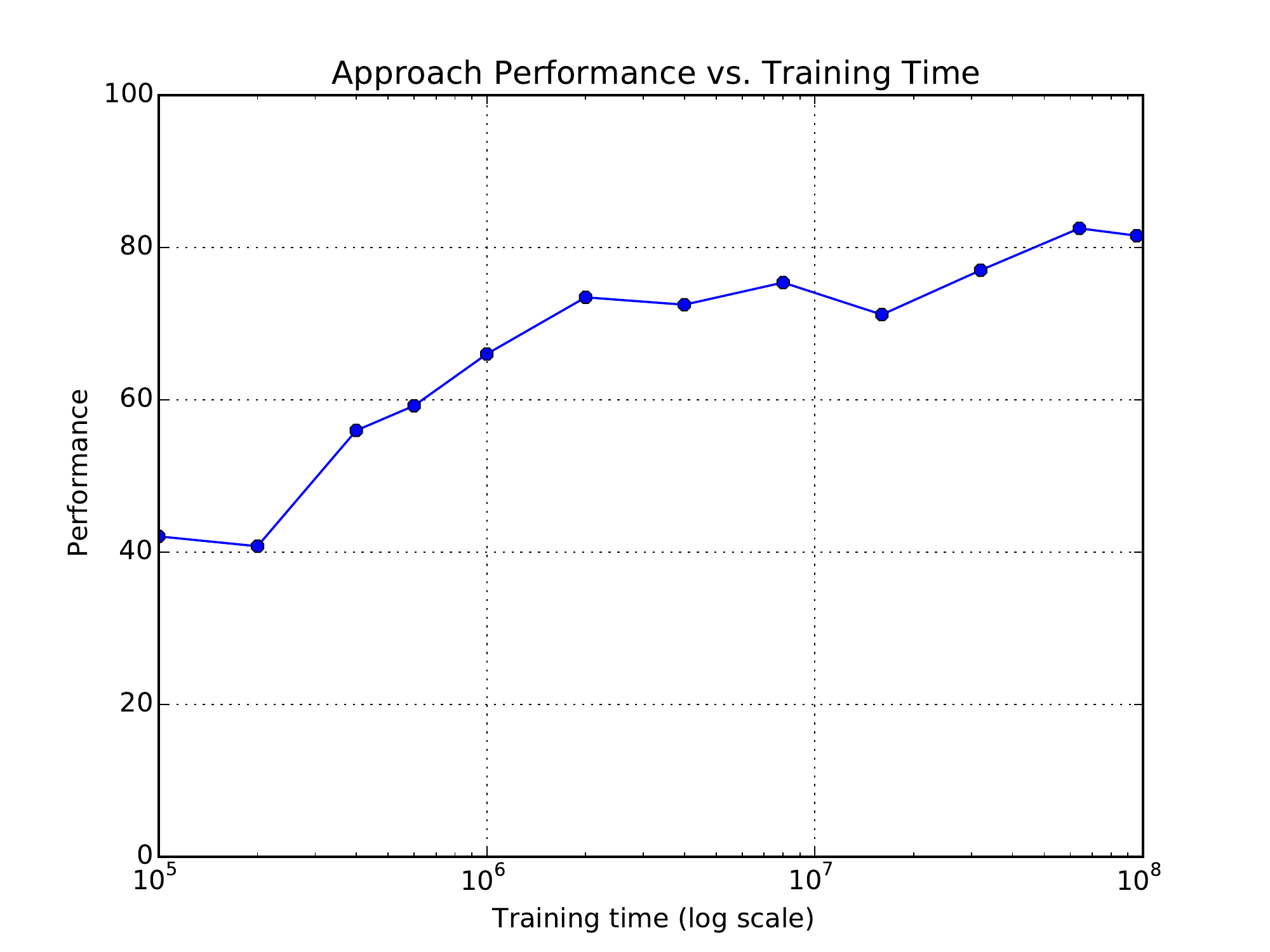}\caption{\label{fig:vw_approach_perf}\textbf{Target Approach Performance in the Virtual World}.   Target approach performance in the virtual world is plotted as a function of the training time of the PVM tracker. Performance is measured as the percentage of trials that the robot reaches the ball before an allotted time-out (see text for details). The performance increases as the PVM tracker is trained longer prior to insertion into the robot.}\end{figure}
Figure \ref{fig:vw_approach_perf} shows a plot of the virtual robot's target approach performance as a function of prior PVM training time. Each data point of the plot is the average performance for 309 trials. The figure shows that performance improves with increasing PVM training time.

\subsection{Experiment 3: Discussion}

We have found that the PVM tracker can be embodied and used on a robot for a task in a realistic virtual world.  This demonstrates that the PVM tracker can be used in an on-line behavior, i.e., in tasks where the motion of the robot and camera are affected by the tracker's own output.

\begin{figure}[ht]\centering\includegraphics[width=0.9\textwidth,clip]{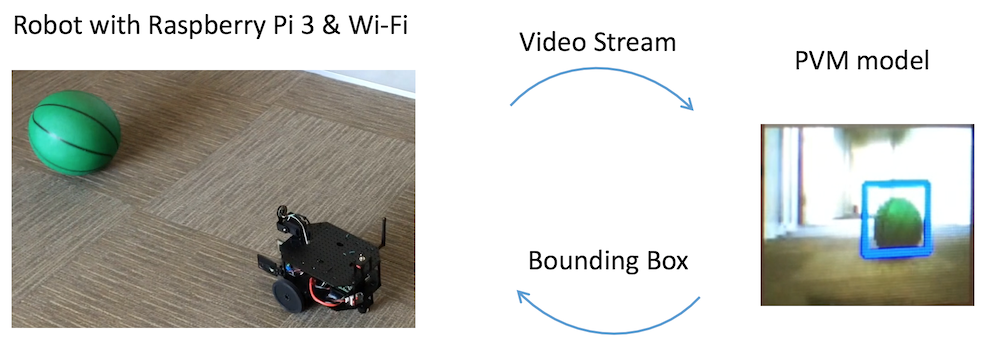}
\caption{\label{fig:real_world_robot_pvm}\textbf{Real-World Test of PVM on a Robot}.  The robot streams video to a PVM tracker. It receives a bounding box for the green ball from the PVM tracker and modifies its motor commands to move towards the ball.}
\end{figure}

A follow-up test checked whether PVM might also work on a physical robot in the real world. We replicated a scenario similar to our virtual world experiments in our office. We linked an instance of PVM to a real robot via Wifi. We found that the PVM was able to track the green ball robustly, as in the virtual world, and the robot was able to approach the basketball successfully even as it moved through varying lighting conditions and shadows.  Figure \ref{fig:real_world_robot_pvm} shows the setup for testing the PVM tracker in the real world.

Although the PVM tracker was trained on hand-held video and under conditions that were significantly different from the virtual world and from the real world as viewed by the robot, performance was good in both cases.  Furthermore, target approach performance improved as the function of increasing training time of the PVM tracker.  This supports the hypothesis that the PVM tracker is able to generalize well to new conditions.

\section{General Discussion}
\label{sec:discussion}
There is great interest in using visual sensors like cameras in many domains because they provide rich data, can function under a wide variety of conditions, and are passive. Algorithms that can learn about the dynamics of the world without human supervision are an important step towards building autonomous systems.  Artificial vision systems capable of such common-sense visual perception would be tremendously valuable in many applications.  

We have shown a new architecture and a methodology that are especially geared towards processing data from visual sensors.  Many in the fields of robotics and machine vision have begun to converge on the idea that machine learning will supplant hand-crafted task-specific algorithms. The recent hope has been that current deep learning methods will approach general real world practicality through the use of more and more data to overcome corner cases.  However, deep learning systems as currently designed have fundamental architectural limitations that make this unlikely. PVM addresses these issues and shows promising results on an online visual object tracking task --- a task which has many advantageous properties, as discussed previously in the paper. Additionally this task can be gradually increased made more complex, ultimately reaching robust closed-loop behavior in physical reality.

The \emph{common sense knowledge} discussed in this paper is composed of a great number of regularities and relations. In order to extract these regularities, vision systems must be designed so they can take into account spatial and temporal context, operate at multiple scales, and have a robust training methodology. Investigators have begun postulating online prediction as a emerging paradigm for training machine learning architectures, and we discuss some of this related work in Section~\ref{sec:comparison}. Since capturing common sense knowledge is likely to require vast quantities of parameters, the system needs to be designed with scalability in mind. In subsection \ref{sec:scalability} we discuss in detail how PVM is designed to be scalable.  In Section~\ref{sec:alternative-context-interpretation} we offer an alternative presentation of how context is used within the PVM.  According to this view, PVM is a self-similar nested architecture. Interpreting PVM in this way is helpful in understanding how it has the property of being both scalable and predictive.

\subsection{Interpreting recurrent feedback in PVM}
\label{sec:alternative-context-interpretation}
The PVM tracker shows impressive results in challenging visual conditions. This includes the ability to track objects even when they are partially obscured or rendered almost invisible, e.g., by harsh backlighting or lens flares (see Figure~\ref{fig:examples}).  This ability suggests that the network has learned to maintain a robust sense of the position of the objects across time and space, despite interfering visual inputs. In this section we discuss how PVM units might generate a persistent memory trace for an obscured object.

\begin{figure}[ht!]
\centering
\includegraphics[width=1.0\textwidth,trim={0cm 9cm 0 0cm},clip]{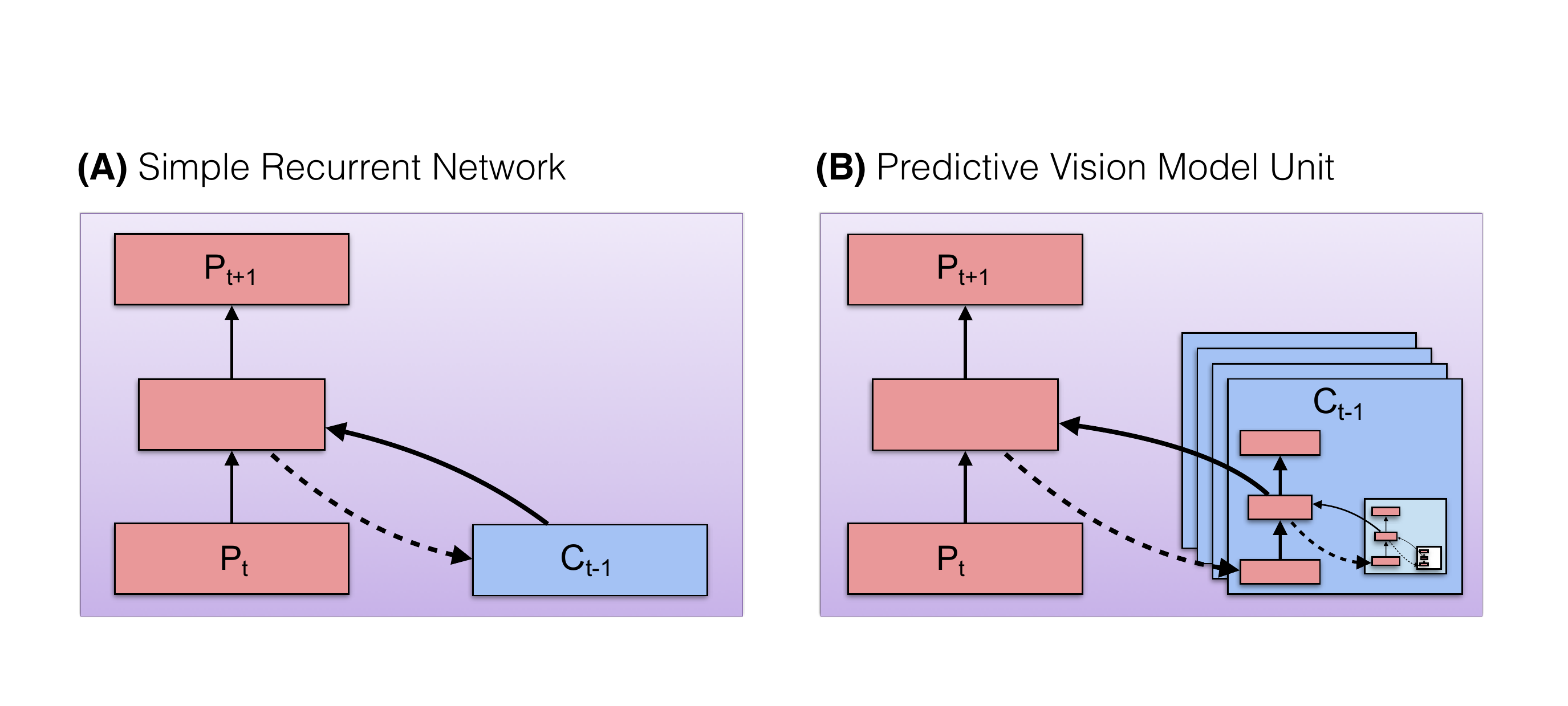}
\caption{\label{fig:pvm_hyperdimensional} \textbf{Analysis of PVM as Nested Recurrent Networks.} (A) A Simple Recurrent Network (SRN) can be configured to predict a future signal $P_{t+1}$ given the current signal $P_{t}$ and prior context, $C_{t-1}$.  The prior context is a copy of the hidden layer activations from the previous time-step. (B) PVM unit re-drawn in terms of an SRN (figure omits the additional signal and the locally computed derivative, integral, and prediction error features). Units organized in a PVM architecture use \emph{each other} as their recurrent context.  This self-similar, nested SRN structure allows context information across both space and time at multiple scales to be brought to each unit for local prediction.  This allows local predictions to take advantage of diffuse as well as local information in a scalable way. PVM can be thought of as a ``hyper-dimensional'' scalable version of an SRN.}
\end{figure}
\FloatBarrier
Some insight into this apparent ``local working memory'' ability can be obtained by re-analyzing the network in terms of the recurrent connections \emph{among} PVM units.  Although PVM is most readily conceived as a hierarchy of units (as shown in Figures~\ref{fig:pvm_hierarchy} and \ref{fig:pvm_circuit}), focusing on recurrent connectivity reveals that PVM units can be seen as a series of spatially and temporally nested Simple Recurrent Networks \citep[SRNs or Elman networks,][]{elman1990finding}.  The PVM replaces the ``context layer'' from SRNs with recurrent inputs from multiple PVM units, as is illustrated in Figure~\ref{fig:pvm_hyperdimensional}. 

SRNs are known to have the ability to maintain information over time to reduce their future prediction errors \citep{elman1990finding}\footnote{SRNs can achieve these predictions across time even when trained in a purely online manner, without optimizations like backpropagation through time \citep{werbos1988generalization}.}. Because PVM units are driven to locally reduce their error, each unit will learn weights to maximally use its lateral and feedback context inputs for predicting the reappearance of an object, even if that object is momentarily or partially obscured from its own primary signal input.  On-going studies in our labs suggest that PVM can predict the reappearance of objects across extended periods of complete occlusion.

A critical difference between SRNs and the PVM units is that whereas SRNs use their own hidden activations as context, PVM units additionally use hidden activations from other PVM units as context. This property may help address the problem of the vanishing gradient, which is also frequently encountered with SRNs. In the case of SRNs, this problem arises because of feedback through nonlinear, squashing activation functions.  This problem is avoided in Long Short-Term Memory units \citep[i.e., LSTMs,][]{hochreiter1997long}, which have internally gated linear feedback, and can also be address through a modification of SRNs \citep{mikolov2014learning}. In the case of PVM, each PVM unit receives context coming from different parts of the PVM network, and consequently has access to information at multiple temporal and spatial scales. As a result, a PVM unit does not rely entirely on its own local recurrent connectivity to support its memory function, reducing the amount of information that has to be circulated repeatedly through the squashing nonlinearities within the unit.  The distributed context may be sufficient to combat the vanishing gradient problem.

\subsection{Scalability}
\label{sec:scalability}

Scalability is a broad term which encompasses multiple properties depending on the exact context. Typically the scalability of an algorithm is judged by the locality of computation versus the necessity to perform synchronous operations that depend on coherent global state. Global synchronizing events typically limit parallelism and contribute to the sequential part of processing described in Amdahl's law \citep{amdahl1967validity}. Within the more specific context of machine learning as applied to robotics, other aspects of scaling come into play: 
\begin{packed_itemize}
\item convergence of the learning procedure --- scalable means that the training time does not diverge with the number of parameters of the model; 
\item real time performance --- if the systems are applied to a robotic system in real world, then the processing time associated with a behavior must not exceed certain constraints imposed by the real world; 
\item latency --- if the systems are used for real-time control, parts of the control task may need to be performed with minimal latency, hence different parts of the system may operate with different real-time requirements (much like biological nervous systems)
\item local computation and power efficiency --- to fulfill real time and latency requirements (as well as robustness to communication interruptions) the systems may need to reside locally on a robot (not in the cloud), which imposes power usage constraints.
\end{packed_itemize}

PVM addresses these scalability challenges in the following ways. The challenge of convergence is addressed by the fact that the training signal is strong at every PVM unit. This is because, unlike in many other algorithms, the training signal (prediction) originates at the unit itself and does need to be propagated through many layers of a deep system. The result is fast convergence at every level of the PVM hierarchy, which based on our initial observations scales linearly or even sublinearly with the size of the system. 

PVM is amenable to implementation on conventional parallel computing hardware --- and perhaps future generations of custom (e.g., neuromorphic) hardware --- due to its uniformity, local learning and mostly local connectivity. Even though substantial amounts of data are moved throughout the system, the data flow pathways are clearly defined and are therefore amenable to hardware acceleration.

PVM design comes with a form of robustness and economy: consistent corruption of the signal can be ``predicted out'' by early stages of the system and does not propagate all the way up the hierarchy and consume resources there\footnote{For example, assume that half of the pixels in the primary input signal of a given PVM unit are always black. The network inside the PVM unit will ultimately learn to predict those constant values using weights from the bias unit in the middle layer. Consequently the internal representation that is output from that unit (its hidden/middle layer) does not need to contain any information about the constant parts of the signal and can remain fully dedicated to representing the part of the signal that actually varies.}. Such robustness tolerates a limited amount of both spatial and temporal noise and allows for weakly synchronized implementations in which every PVM unit runs on a separate real-time clock. Such implementations can overcome timing bottlenecks which affect large systems due to Amdahl's law \citep{amdahl1967validity}.  Consistent clock discrepancies become part of the predicted reality for downstream units and can be ``predicted out'' by the system.  

\subsection{Comparison to related approaches}
\label{sec:comparison}
The work presented here is substantially inspired by the Hierarchical Temporal Memory (HTM) \citep{hawkins2007intelligence} and we subscribe to many ideas presented there. At the same time the complexity of biology and often contradicting findings in neuroscience provide insufficient constraints to design HTM-like structures from the single neuron level. Instead, we opted to explore more abstract ideas using well understood components from machine learning. 

An important class of related vision models called Neural Abstraction Pyramid were proposed in \citep{behnke1998neural} and expanded upon in several subsequent publications e.g. \citep{behnke1999hebbian, behnke2003hierarchical}. Contrary to many more recent approaches, the Neural Abstraction Pyramid proposed feedback and lateral connectivity as efficient mechanism to include contextual information for resolving local ambiguities in image interpretation. Competitive unsupervised learning of hierarchical representations and end to end backpropagation training of multiple computer vision tasks like image reconstruction and face localization were demonstrated with this architecture \citep{behnke2003hierarchical}. The vanishing gradient problem was addressed with RProp \citep{riedmiller1993direct}. Even though the network was trained only with stationary targets, the tracking of moving faces was demonstrated in \citep{behnke2002learning}.

Recently several papers have appeared postulating prediction of video frames as a paradigm for unsupervised training  \citep[][to cite a few]{palm2012prediction, vondrickanticipating,ranzato2014video,lotter2015unsupervised}. All of these approaches rely on convolutional nets as the underlying learning architecture. Although their use of convolutional nets means these models inherit many good properties (e.g., good performance on some benchmarks and efficient GPU implementations), it also means they inherit many undesirable properties (e.g., the loss of important region-specific and spatial information, the inability to use context or feedback). In models that use convolutional net as a perceptual front-ends, there have been several efforts to address issues of time and context. Some approaches use LSTM layers \citep{hochreiter1998vanishing} during later stages of processing at the higher, more abstract layers of the networks. However, LSTM cannot be directly used in convolutional or pooling layers, and feedback in LSTM networks is restricted to within the layer of LSTM units \citep[e.g.,][]{xiong2016dynamic}. In other similar cases where convolutional nets are used as a perceptual front-end \citep{yang2015robot}, the dynamical relations derived from that front-end are processed using a grammar based model.  In all of these cases, prediction and perception are separated, precluding the inference of dynamics at multiple scales.

In another recent attempt, layers of two convolutional nets were ``cross stitched'' for re-usability of features in multitask learning \citep{misra2016cross}. The idea resembles lateral context in PVM, however PVM applies context in a more widespread manner. 

The importance of spatial context in visual learning was explored in work by \cite{doersch2015unsupervised}. In this case a convolutional network was successfully trained to predict a relative arrangement of two image patches --- confirming that visual data contain rich information in the form of spatial context.

One approach that has highly similar motivation to ours can be found in \cite{finn2016unsupervised}.  Like our work, their model also departs from the primarily feedforward approach seen in much of deep learning and introduces recurrent layers early on in processing.  Also similarly, their model feeds the state of the robotic arm as a form of context back into one of its layers. The architecture presented there corresponds most directly to a single one of our PVM units with a context input (where the context is motor context rather than visual context), and is much deeper and more complex than our PVM unit. Two critical differences in our approach are that (1) we opted to use large number of simpler units, and (2) we moved away the end-to-end training paradigm (to overcome scalability and generalization issues discussed above). 

\section{Conclusions}
\label{sec:conclusions}
\subsection{Summary}

We have introduced a new approach to artificial vision which involves a hierarchical, recurrent architecture trained using unsupervised methods. This approach was borne out of our need for systems capable of robust,
stable visual perception in challenging visual conditions while also satisfying the requirements of online learning and scalability. The design of the architecture allows it to take into account temporal and contextual features of the visual world, enabling it to overcome challenges like illumination, specular reflections, shadows, and momentary occlusions.  We assessed our unsupervised system by attaching a supervised visual object tracking task and found that it could visually track objects at performance levels on par with or better than expert tracking algorithms. 

The architecture breaks with the prevailing lineage of Neocognitron-inspired systems \citep{fukushima1980neocognitron} and incorporates many ideas gathered over the past 50 years in neuroscience. Prediction as a paradigm for training is compatible with a broader view of intelligent behavior in physical environments, based on causal entropy \citep{wissner2013causal}. We show encouraging results accomplished with PVM for visual object tracking and conjecture that similar architectures may be useful in a multitude of robotic/perception tasks. 

The central research problem driving the research presented in this paper is Moravec's paradox \citep{moravec1988mind}: the question of real world perception and mobility.
Although some accomplishments have been achieved in applying machine learning to robotics \citep{levine2016learning}, we are still far from a robot that would be able to robustly roam around an average human household and manipulate objects (say, pick toys up from the floor). In our view, a robust scalable machine learning architecture that spontaneously learns a model of physical reality is a key ingredient in building truly intelligent robots\footnote{Clearly this is not the only needed ingredient; much as the neocortex is not the only part of the brain, many other functions are likely needed, including task selection, reward/reinforcement, reflexes, planning, among others.}.  

We further contend that an understanding of reality obtained in this way is a prerequisite even for non-embodied AI systems e.g., high-functioning personal intelligent assistants, or more accurate language translation systems. Once common-sense knowledge has been learned in embodied systems, it may then be transferable to these kinds of non-embodied systems, permitting them to have improved referential understanding.  In any case, embodied and non-embodied AI systems will need to connect to a human-centered world. 
Although efforts like ImageNet do attempt to make this connection between AI and the human-centered world by learning direct mappings between visual images and human-centered category labels, it is premature to expect that such systems will generalize correctly. Instead, it seems to us that robust generalizing AI systems must be developed, grown, and matured starting from basic, early perceptual (and possibly sensori-motor) components learned via their immersion in the real world.

\subsection{Future research}
PVM is a new architecture that shows promise for learning the \emph{common sense} knowledge necessary for robust robot behavior in unrestricted environments. There are numerous directions in which the model can be investigated:
\begin{packed_itemize}
\item Explore the space of arrangements of PVM units, long range connections, long range feedback
\item Apply PVM in non-visual applications
\item Apply PVM to a cross-modal problem to explore the concept of co-prediction
\item Explore alternative implementations of the PVM unit
\item Explore the intrinsic dynamics of the network
\item Quantify the effects of feedback and lateral connections
\item Leverage the experience of deep learning in building a better PVM
\end{packed_itemize}
In subsections below we discuss in more detail other relevant directions and how PVM could fit into the field of neuromorphic chip design and neuroscience. 

\label{sec:future}
\subsubsection{New tasks}
Visual object tracking is a task of practical importance and a large field on its own, however our real goal is to evaluate our systems on a set of tasks of increasing complexity leading eventually to robust behavior in physical reality. We anticipate progressing from object tracking to target approach; target approach in progressively more complex environment where there are obstacles to avoid (which may obscure the view of the target); target approach requiring landmark navigation and so on. Part of that research could be realized using high quality virtual worlds -- as provided by contemporary game engines with high quality photo-realistic rendering as well as physical simulation, see Section~\ref{section:virtual} for preliminary results using Unreal\texttrademark\  game engine. 

\subsubsection{Path to practical neuromorphic architectures}\label{sec:path_neuromorphic}

As noted in section \ref{sec:unit_details} the essential component for the PVM meta-architecture is a simple associative memory unit. In the present work, this memory unit is implemented using a multilayer perceptron and therefore may not be considered strictly neuromorphic at the component level\footnote{But it captures many other aspects of the brain at the architecture level design.}. However, nothing fundamental stands in the way of replacing the MLP with other algorithms, possibly even algorithms having nothing to do with neural networks. The implementations can naturally progress from perhaps less power efficient, but relatively easy to fabricate digital vector units (e.g., GPUs) towards more custom digital ASICs (e.g., \mbox{SpiNNaker} \cite{furber2014spinnaker}, TrueNorth \cite{cassidy2014real}) all the way to very power efficient analog circuits operating with artificial spiking neurons (e.g., \cite{schemmel2012live}, \cite{benjamin2014neurogrid}).  This gives a lot of freedom in designing a neuromorphic architecture. We believe that the design of neuromorphic hardware has to accompany the continuation of the research presented here, since likely only much larger instances than what is presented currently (likely by 2-3 orders of magnitude\footnote{This is a guess based roughly on numbers for primate brains.}) will start showing capabilities sufficient for robust behavior in unconstrained environments. Current models already take weeks to train on top of the line CPUs and could be accelerated by an order of magnitude on GPUs, but real-time operation and power requirements will necessitate neuromorphic solutions in the longer term.

\subsubsection{Human machine interaction and embodiment}
As we've mentioned in sections above, we ultimately want to get away from the dataset-based approach and move towards embodiment of PVMs in dynamical environments, both virtual and real. Once a robot is placed in a real world setting, a number of questions arise:
\begin{packed_itemize}
\item How can one reliably and naturally communicate a task to the system --- whether it be to track an object, follow an object, or move to a location? 
\item Is there a natural way to apply reinforcement learning to a system based on PVM?
\item Could there be some hybrid approaches with algorithmic logic on top of perceptual primitives learned by PVM (similar to \cite{yang2015robot})?
\item How can one use PVM to learn to perform multiple tasks, where each task may result in different distinct kinds of predictions being made?
\item How can a primarily unsupervised machine interact well with humans? And is there any way to guarantee safety of a largely unsupervised or self-supervised system while interacting with humans?
\end{packed_itemize}

Many of these robotics questions have parallels within the context of neuroscience/cognitive science. For example, questions about action selection and decision-making parallel neuroscientific questions about the function of non-cortical regions like basal ganglia in processing the signal from cortex, and how the reward system works. As much as neuroscience may not be the easiest way to get these answers (due to the inherent complexity of the brain, and the difficulty in isolating the relevant effects), some convergence of the fields might yield fruitful insights.

\subsubsection{Reinforcing the bridge between theory and neuroscience}
Neuroscience generates large amounts of data, and computational efforts are producing extremely detailed and large simulations of small circuits of neurons  \citep[e.g.,][]{markram2015reconstruction}. However, theory for understanding brain function is lagging -- particularly when it comes to how their modes of operating and capabilities are established. Brains are complex biological systems, and it remains unclear which details are crucial to the actual computation being performed.  Given the complexity and limited time and resources, it is imperative to build good theories to allow us to sieve out the massive amount of data and conduct only those experiments informed by theory. 

The work presented here attempts to bring several disparate communities closer: machine learning researchers, neuromorphic engineers, the neuroscience community and developmental roboticists. Finding some common ground, in our view, will be pivotal to future progress in all of these fields.  PVM incorporates multiple ideas discussed by cognitive neuroscience researchers including brains as predictive systems \citep{clark2013whatever}, elements of predictive coding \citep{rao1999predictive}, and the feedback and lateral connectivity missing from the Neocognitron \citep{fukushima1980neocognitron}. PVM implements these concepts using a well-known machine learning building block (the multi-layer perceptron), while leaving open the possibility of adapting the same idea with more biologically-plausible models and neuromorphic implementations. Studies of PVM variants implemented with e.g., spiking associative networks as their units may lead to insight about how to interpret what we see in the neocortex, and invigorate theory-motivated efforts in neuroscience. Finally PVM addresses several issues raised by developmental roboticists (see particularly sections III and IV in \cite{sigaud2016towards}) in hope to allow robots the represent their environment and affordances. 

\subsubsection{Source code}

We note that the ubiquitous feedback connectivity in PVM makes it challenging to implement in any of the numerous existing deep learning frameworks, such as Torch \citep{collobert2002torch}, TensorFlow \citep{abadi2016tensorflow}, Theano \citep{bergstra2010theano}, and Caffe \citep{jia2014caffe}, among others. 
Therefore, we release source code for our own multicore implementation of PVM to allow other researchers to reproduce our results and experiment with their own ideas. The implementation is available at \url{http://www.github.com/braincorp/PVM}.

\subsection{Acknowledgments}
The majority of the results presented here were accomplished with the support of the Air Force Research Laboratory (AFRL) and DARPA Cortical Processor Seedling Grant under contract FA8750-15-C-1078. We acknowledge Dan Hammerstrom who started the program. We would also like to acknowledge the many brave robots developed in our labs.  They helped us to see what happens when beautiful abstract ideas collide with ugly physical reality --- and how the most basic human perceptual mechanisms are still marvelous compared to anything in Machine Learning. 

\bibliographystyle{apalike}
\bibliography{pvm.bib}

\end{document}